\documentclass[final]{cvpr}

\usepackage{times}
\usepackage{epsfig}
\usepackage{graphicx}
\graphicspath{ {./figs/} }
\usepackage{amsmath}
\usepackage{amssymb}

\usepackage[toc,page]{appendix}
\usepackage{algorithm}
\usepackage{algorithmic}
\usepackage{enumitem}
\usepackage[flushleft]{threeparttable} 
\usepackage{multirow}
\usepackage{subfigure}

\usepackage[pagebackref=true,breaklinks=true,colorlinks,bookmarks=false]{hyperref}

\def\ie{i.e.}
\def\eg{e.g.}

\def\and{\textrm{and}}

\def\0{\textbf{0}}
\def\1{\textbf{1}}

\def\v{\boldsymbol{v}}

\def\q{\boldsymbol{q}}

\def\x{\boldsymbol{x}}

\def\u{\boldsymbol{u}}

\def\L{\mathcal{L}}

\def\T{\mathcal{T}}

\newcommand{\RR}{I\!\!R} 

\newcommand{\myparagraph}[1]{\noindent\textbf{#1.}}

\begin{document}


\title{Learning a Self-Expressive Network for Scalable Subspace Clustering}
\title{Learning an Elastic Self-Expressive Network for Scalable Subspace Clustering}
\title{Learning an Elastic Self-Expressive Network for Subspace Clustering on Infinite Sample}
\title{Learning an Elastic Self-Expressive Network for Scalable Subspace Clustering on Infinite Sample}
\title{Learning an Elastic Self-Expressive Network for Subspace Clustering}
\title{Learning Self-Expression Network for Subspace Clustering}
\title{Learning a Self-Expressive Network for Subspace Clustering}

\author{First Author\\
Institution1\\
Institution1 address\\
{\tt\small firstauthor@i1.org}
\and
Second Author\\
Institution2\\
First line of institution2 address\\
{\tt\small secondauthor@i2.org}
}

\author{Shangzhi~Zhang$^\dag$,~Chong~You$^\ddag$,~Ren\'{e}~Vidal$^\S$~and~Chun-Guang Li$^\dag$\\
$^\dag$ School of Artificial Intelligence, Beijing University of Posts and Telecommunications \\
$^\ddag$ Department of EECS, University of California, Berkeley, CA\\  
$^\S$ Mathematical Institute for Data Science, Johns Hopkins University, Baltimore, MD
}

\maketitle

\begin{abstract}

State-of-the-art subspace clustering methods are based on the self-expressive model, which represents each data point as a linear combination of other data points. However, such methods are designed for a finite sample dataset and lack the ability to generalize to out-of-sample data. Moreover, since the number of self-expressive coefficients grows quadratically with the number of data points, their ability to handle large-scale datasets is often limited. In this paper, we propose a novel framework for subspace clustering, termed Self-Expressive Network (SENet), which employs a properly designed neural network to learn a self-expressive representation of the data. We show that our SENet can not only learn the self-expressive coefficients with desired properties on the training data, but also handle out-of-sample data. Besides, we show that SENet can also be leveraged to perform subspace clustering on large-scale datasets. Extensive experiments conducted on synthetic data and real world benchmark data validate the effectiveness of the proposed method. In particular, SENet yields highly competitive performance on MNIST, Fashion MNIST and Extended MNIST and state-of-the-art performance on CIFAR-10. The code is available at: \href{https://github.com/zhangsz1998/Self-Expressive-Network}{https://github.com/zhangsz1998/Self-Expressive-Network}




\end{abstract}

\section{Introduction}
\label{sec:intro}

With technological advances in data acquisition, storage and processing, there is a surge in the availability of \emph{large-scale} databases in computer vision. While the development of modern machine learning techniques, such as deep learning,
has led to great success in analyzing big data, such methods require a large amount of annotated data which is often costly to obtain. Extracting patterns and clusters from \emph{unlabeled} big data has become an important open problem.

We consider the problem of clustering large-scale unlabeled data under the assumption that each cluster is approximated by a low-dimensional subspace of the high-dimensional ambient space, a.k.a. \emph{subspace clustering}~\cite{Vidal:SPM11-SC, Vidal:Springer16}. This problem has wide applications in image clustering \cite{Ho:CVPR03, Elhamifar:TPAMI13}, motion segmentation~\cite{Costeira:IJCV98, Chen:IJCV09}, hybrid system identification~\cite{Vidal:ACC04, Bako-Vidal:HSCC08}, cancer subtype clustering~\cite{McWilliams:DMKD14,Li:TIP17}, hyperspectral image segmentation \cite{Zhang:TGRS16} and so on.

\emph{Self-expressive} model~\cite{Elhamifar:CVPR09} is one of the most popular and successful methods for subspace clustering. Given a data matrix $X = [\x_1, \cdots, \x_N] \in \RR^{D\times N}$ whose columns are drawn from a union of $n$ subspaces, the self-expressive model expresses each data point $\x_j \in \RR^{D}$ as a linear combination of other data points, \ie,
\begin{align}
\label{eq:self-expression}
\x_j =\sum_{i\neq j} c_{ij} \x_i, 
\end{align}
where $\{c_{ij}\}_{i \neq j}$ are self-expressive coefficients.
A remarkable property of the self-expressive model is that solutions to \eqref{eq:self-expression} that minimize certain regularization function on the coefficients have the \emph{subspace-preserving property}, i.e., nonzero coefficients $c_{ij}$ occur only between $\x_i$ and $\x_j$ lying in the same subspace \cite{Elhamifar:CVPR09,Elhamifar:TPAMI13,Liu:ICML10,Lu:ECCV12,Soltanolkotabi:AS12,Wang:NIPS13-LRR+SSC,You:ICML15,You:CVPR16-EnSC,Yang:ECCV16,Lu:TPAMI18}. Consequently, correct clustering can be obtained by defining an affinity between any pair of data points $\x_i$ and $\x_j$ as, \eg, $|c_{ij}|+|c_{ji}|$, and applying spectral clustering to the affinity. Recent developments further extend the applicability of self-expressive models to the case where the data are corrupted by noise \cite{Wang-Xu:ICML13,Soltanolkotabi:AS14,Wang:JMLR16} and outliers \cite{Soltanolkotabi:AS12,You:CVPR17}, are imbalanced over classes \cite{You:ECCV18}, or possess missing entries \cite{Tsakiris:ICML18}.

Despite its great empirical performance and broad theoretical guarantees for correctness, the self-expressive model suffers from the limitation that it requires solving for a self-expressive matrix of size $N\times N$, which is computationally prohibitive for large-scale data. Although scalable subspace clustering methods based on subsampling~\cite{Peng:CVPR13}, sketching \cite{Traganitis:TSP18} or learning a compact dictionary \cite{Adler:TNNLS15,Shen:ICML16} already exist, they do not have broad theoretical guarantees for correctness and sacrifice accuracy for scalability. 
In addition, the self-expressive coefficients computed for a set of data cannot be used to produce self-expressive coefficients for previously unseen data, posing challenges for learning in an online setting and for 
out-of-sample data.

In this work, we introduce the \emph{self-expressive network (SENet)} to learn a self-expressive model for subspace clustering, which can be leveraged to handle 
out-of-sample data and large-scale data.
Our method is based on learning a function $f(\x_i, \x_j; \Theta): \mathbb{R}^D \times \mathbb{R}^D \to \mathbb{R}$, implemented as a neural network with parameters $\Theta$, that is designed to satisfy the self-expressive model
\begin{equation}
\label{eq:self-expression-function}
    \x_j = \sum_{i \ne j} f(\x_i, \x_j; \Theta) \cdot \x_i.
\end{equation}
In principle, the number of network parameters does not need to scale with the number of points in the dataset, hence SENet can effectively handle large scale data. Moreover, an SENet trained on a certain dataset can be used to produce self-expressive coefficients for another dataset drawn from the same data distribution, therefore the method can handle out-of-sample data effectively.
We present a network architecture for $f(\x_i, \x_j; \Theta)$ as well as a training algorithm that allow us to learn self-expressive coefficients with desired subspace-preserving properties.
Our experiments showcase the effectiveness of our method as summarized below:
\begin{enumerate}[leftmargin=*,topsep=0.25em,noitemsep]
    \item We show that the self-expressive coefficients computed by a trained SENet closely approximate those computed by solving for them directly without the network. This illustrates the ability of SENet to approximate the desired self-expressive coefficients.
    \item We show that a SENet trained on (part of) the training set of MNIST and Fashion MNIST can be used to produce self-expressive coefficients on the test set that give a good clustering performance. This illustrates the ability of SENet to handle out-of-sample data.

    \item We show that SENet can be used to cluster datasets containing $70,\!000$+ data poins, such as  MNIST, Fashion MNIST and Extended MNIST, very efficiently, achieving a performance that closely matches (for MNIST, Fashion MNIST and Extended MNIST) or surpasses (for CIFAR-10) the state of the art.
\end{enumerate}

\section{Related Work}
\label{sec:related-work}


\myparagraph{Deep Clustering}
Our work is fundamentally different from many existing studies on jointly training a deep neural network and learning self-expressive coefficients \cite{Peng:IJCAI16,Ji:NIPS17,Peng:arxiv17,Zhou:CVPR18,Yang:arxiv2019,Zhang:ICML19,Zhang:CVPR19} for subspace clustering.
In such methods, deep networks are used to extract features (so that they lie in linear subspaces) from input data (which may not lie in linear subspaces), and self-expressive model is applied in the feature space \cite{haeffele2021a,abdolali2021beyond}.
In contrast, our work assumes that the input data already lie 
in linear subspaces, and focuses on computing the self-expressive coefficients.
Our work also shares similarities with SpectralNet \cite{Shaham:ICLR18}, which learns a neural network to produce a latent embedding by optimizing a spectral clustering objective on an 
affinity graph.
Such a method does not have a low-dimensional modeling for data therefore is different from ours.\footnote{When finalizing the submission, 
we became aware of a work-in-progress report \cite{busch2020learning} that presents a similar idea as ours.
While \cite{busch2020learning} uses $\ell_2$ regularization and imposes symmetry on self-expressive coefficients, our model uses a general elastic net regularization and does not impose symmetry constraint, therefore has a better capability of obtaining subspace-preserving properties. 
}


\myparagraph{Self-expressive Models}
Many works have explored different choices of regularization on the self-expressive coefficients for subspace clustering.
For instance, $\ell_1$ regularization is used in sparse subspace clustering~\cite{Elhamifar:CVPR09,Elhamifar:TPAMI13}, for which the optimal solution is subspace-preserving when the subspaces are independent, disjoint, intersecting or even affine ~\cite{Elhamifar:TPAMI13,Soltanolkotabi:AS12,Wang-Xu:ICML13,Wang:JMLR16,You:ICML15,Li:JSTSP18, You:ICCV19,robinson2019basis};
nuclear norm and $\ell_2$ norm regularization are used in low-rank~\cite{Liu:ICML10, Favaro:CVPR11} and least squares subspace clustering~\cite{Lu:ECCV12}, respectively, for which the optimal solution is subspace-preserving when the subspaces are independent;
mixing $\ell_1$ norm with either $\ell_2$ or nuclear norm regularization are used in \cite{You:CVPR16-EnSC} and \cite{Wang:NIPS13-LRR+SSC}, respectively, to improve connectivity of affinity graph while maintain broad theoretical guarantees for subspace-preserving property.
In addition, there are works on noise modeling~\cite{Li:CVPR15MoG, Li:CVPR19-subspace, He:TNNLS16} and feature learning~\cite{Liu:ICCV11, Patel:ICCV13, Peng:CVPR17, Ji:NIPS17, Zhou:CVPR18, Zhang:CVPR19, Yu:NIPS20} for self-expressive models. 

\myparagraph{Scalable Subspace Clustering}
Due to its importance in practical applications, large scale subspace clustering has drawn a lot of research attentions.
An early work \cite{Peng:CVPR13} presented a subsampling based approach in which a random subset of data is sampled and clustered, then the rest of the data are classified with sparse representation based classification \cite{Wright:PAMI09}.
Following this work, several methods adopt a two-step approach for computing self-expressive coefficients: 1) construct a dictionary, either generated in random \cite{Traganitis:TSP18} or learned/selected from data \cite{Shen:ICML16,Adler:TNNLS15,You:ECCV18,Aldroubi:ACHA17,abdolali2019scalable,Matsushima:NeurIPS19}, and 2) express each data point as a linear combinations of the atoms in the dictionary.
In particular, motivated by the development of learned optimization solvers such as LISTA~\cite{Gregor:ICML10} and ISTA-Net \cite{Zhang:CVPR18} for solving sparse optimization problems, \cite{Li:AAAI17, Li:arxiv20} presented a framework 
where one jointly solves for the self-expressive coefficients and trains a neural network to approximate self-expressive coefficients with a dictionary in the first step, so that the computation of self-expressive coefficients in the second step can be carried out efficiently. %
In principle, the clustering performance of such a two-step approach increases with the size of the dictionary.
However, the output dimension hence the scale of the optimization problem in \cite{Li:AAAI17, Li:arxiv20} increases at least quadratically with the size of the dictionary, therefore using a sufficiently large dictionary may be impossible.

Another group of methods achieve efficient computation by decomposing a large-scale optimization problem into a sequence of small scale problems, by either a greedy approach~\cite{You:CVPR16-SSCOMP,Dyer:JMLR13}, active support method~\cite{You:CVPR16-EnSC}, or dropout strategy~\cite{Chen:CVPR20}. These methods enjoy broad theoretical guarantees for correctness and have superior empirical performance. Nonetheless, they have quadratic time and memory requirement, therefore cannot handle very large scale data.

\myparagraph{Self-attention Models}
The self-attention mechanism used in Graph Attention Networks (GAT) \cite{Velick:ICLR18}, Transformer~\cite{Vaswani:NIPS17}, Non-local Neural Networks \cite{Wang:CVPR18}, etc., shares similar idea with the self-expressive models.
In these works, the (output) features of one data point are computed as a linear combination of (input) features of all data points.
Similar to SENet, the coefficients in the linear combination are computed with a neural network.
However, unlike the self-expressive models, which use the distance between the input features and output features to define a training loss in an unsupervised manner, the self-attention methods impose a supervised learning loss on the output features.
This leads to a difference in the design of the network architecture, as we explain in the next section.

\section{Self-Expressive Network}

\subsection{Model}
\label{subsec:model}

Let $X = [\x_1, \cdots, \x_N] \in \RR^{D\times N}$ be a data matrix whose columns lie in a union of low-dimensional linear subspaces of $\RR^D$. Self-expressive methods for subspace clustering are based on solving for every $j \in \{1, \cdots, N\}$ an optimization problem of the form
\begin{align}
\min_{\{c_{ij}\}_{i \ne j}} \frac{\gamma}{2}\| \x_j -   \sum_{i \ne j}c_{ij} \x_i \|_2^2 + \sum_{i \ne j} r(c_{ij}),
\label{eq:self-expression-optimization}
\end{align}
where $r(\cdot): \RR \mapsto \RR_+$ is a regularization function and $\gamma>0$ is a balancing parameter.
The idea is that any column $\x_j$ can be expressed as a linear combination of other columns of $X$ that are \emph{from the same subspace as $\x_j$}.
Such a linear combination is known as \emph{subspace-preserving}, and it can be recovered by solving \eqref{eq:self-expression-optimization} with certain choices of regularization $r(\cdot)$.
Aggregating the solutions to \eqref{eq:self-expression-optimization} for all columns of $X$ 
yields a self-expressive coefficient matrix $C \in \RR^{N\times N}$ with the $i,j$-th entry given by $c_{ij}$. 
When $C$ is subspace-preserving, spectral clustering \cite{vonLuxburg:StatComp07} on an affinity given by, e.g., $|C| + |C^\top|$, produces correct clustering of the data matrix $X$.

We present a method that is based on solving the following optimization problem in lieu of \eqref{eq:self-expression-optimization}:
\begin{align}
\begin{split}
\label{eq:self-expression-functional-optimization}
\min_{\Theta} \frac{\gamma}{2}\| \x_j - \sum_{i\ne j} f(\x_i, \x_j; \Theta) \x_i \|_2^2 + \sum_{i \ne j} r\big(f(\x_i, \x_j; \Theta)\big),\!\!\!\!\!
\end{split}
\end{align}
%
where $f(\x_i, \x_j; \Theta): \mathbb{R}^D \times \mathbb{R}^D \to \mathbb{R}$ is a function parameterized by $\Theta$.
There are two benefits of using the model in \eqref{eq:self-expression-functional-optimization} over the model in \eqref{eq:self-expression-optimization}.

First, the number of parameters in \eqref{eq:self-expression-optimization} (collectively for all $j \in \{1, \ldots, N\}$) is quadratic with the number of data points $N$, which limits its applicability to large scale datasets since an $N$-by-$N$ matrix may not fit into memory.
In contrast, the number of parameters in \eqref{eq:self-expression-functional-optimization} needs not be related to the number of data points, and can be determined flexibly based on the availability of the memory.
In principle, the model in \eqref{eq:self-expression-functional-optimization} 
may be used to compute self-expressive coefficients for datasets of arbitrary size.

Second, self-expressive coefficients computed from \eqref{eq:self-expression-optimization} for a particular dataset cannot be used for another dataset that is drawn from the same distribution.
This implies that the model in \eqref{eq:self-expression-optimization} cannot be used to handle out-of-sample data, for which self-expressive coefficients need to be computed from scratch.
In contrast, a self-expressive function in \eqref{eq:self-expression-functional-optimization} once learned on a particular dataset can be used to generate self-expressive coefficients for out-of-sample data.
By our design 
of the network architecture for $f(\cdot, \cdot; \Theta)$ as we discuss in Subsection~\ref{subsec:architecture}, the calculation on out-of-sample data can be carried out very efficiently.

\myparagraph{Choice of Regularization $r(\cdot)$}
It is known that sparsity regularization in self-expressive models enforces subspace-preserving properties under broadest conditions~\cite{Elhamifar:CVPR09,Elhamifar:TPAMI13, Soltanolkotabi:AS12, You:ICML15, Wang:JMLR16, Tsakiris:ICML18, Li:JSTSP18, You:ICCV19}. For example, the work \cite{Soltanolkotabi:AS12,Wang:JMLR16} showed that with $\ell_1$ regularization on the coefficients, the model in \eqref{eq:self-expression-optimization} produces subspace-preserving solutions even when the subspaces intersect, provided that the subspaces are sufficiently separated and points in each subspace are 
well-distributed.
On the other hand, sparsity regularization produces solutions that have too many false negatives, \ie, 
the self-expressive coefficient $c_{ij}$ can often be zero even when $\x_i$ and $\x_j$ are from the same subspace.
This may lead to a poorly connected affinity graph that results in over-segmentation.
Hence, the work \cite{You:CVPR16-EnSC} advocated using elastic net regularization, which is given by a weighted sum of $\ell_1$ and $\ell_2^2$ regularization with a balancing parameter $\lambda \in [0, 1]$:
\begin{equation}
\label{eq:elastic-net-regularization}
    r(\cdot) = \lambda |\cdot| + \frac{1-\lambda}{2} (\cdot)^2.
\end{equation}
This regularizer provably produces subspace-preserving solutions under similar conditions as for the $\ell_1$ regularizer, and at the same time produces a denser coefficient matrix, hence an improved clustering performance. Therefore, 
we adopt elastic net regularization for our model in \eqref{eq:self-expression-functional-optimization}.

\subsection{Network Instantiation}
\label{subsec:architecture}

Inspired by recent advances in deep learning, we implement the self-expressive function $f(\cdot,\cdot;\Theta)$ in our model \eqref{eq:self-expression-functional-optimization} via a deep neural network with training parameters $\Theta$.
We refer to the network as 
\emph{Self-Expressive Network} \textbf{(SENet)}.

\begin{figure}
\begin{center}
\includegraphics[scale=0.375]{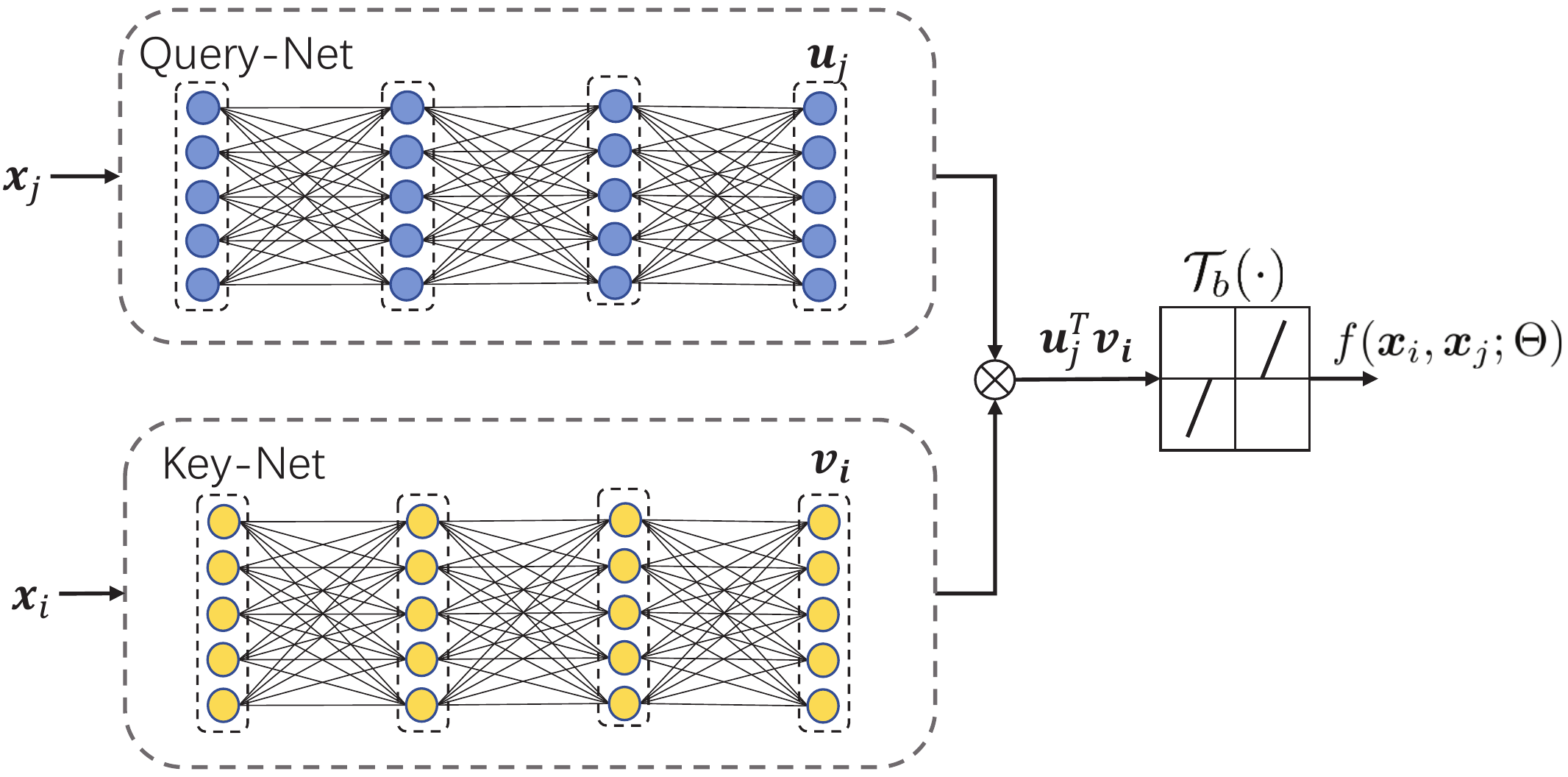} 
\end{center}
\caption{Architecture of Our SENet} 
\label{fig:architecture-SENet}
\end{figure}

Specifically, we propose the following network formulation for SENet: 
\begin{align}
\begin{split}
f(\x_i, \x_j;\Theta) = \alpha \T_b(\u^\top_j \v_i),
\end{split}
\label{eq:SENet-formulation}
\end{align}
where
\begin{align}
               \u_j &:= \u(\x_j;\Theta_u) \in \RR^p,\label{eq:SENet-formulation-u}\\
               \v_i &:= \v(\x_i;\Theta_v) \in \RR^p.\label{eq:SENet-formulation-v}
\end{align}
%
In above, $\u(\cdot; \Theta_u)$ and $\v(\cdot;\Theta_v)$, referred to as query and key networks, are two multilayer preceptrons (MLPs) that perform mappings $\RR^D \mapsto \RR^p$ with learnable parameters $\Theta_u$ and $\Theta_v$, respectively, where $p$ is a model 
parameter. $\T_b(\cdot)$ is a learnable soft thresholding operator defined as
\begin{align}
\T_b(t) := \text{sgn}(t) \max(0,|t|-b),
\label{eq:learnable-soft-threshold}
\end{align}
where $b$ is a learnable parameter and $\alpha > 0$ is a fixed numerical constant.
For clarity, we use $\Theta := \{ \Theta_u, \Theta_v, b \}$ to denote all the trainable parameters in SENet, and illustrate the architecture of the neural network $f(\x_i, \x_j;\Theta)$ 
in Fig.~\ref{fig:architecture-SENet}.

By the design of network architecture in \eqref{eq:SENet-formulation}, the self-expressive coefficient for a pair of data points $(\x_j, \x_i)$ is computed by learning a pair of representations 
$\u_j$ and $\v_i$, respectively, and taking the inner product of $\u_j$ and $\v_i$ before applying a soft-thresholding.
We empirically find (see Sec.~\ref{sec:Synthetic Experiments}) that such a network can produce self-expressive coefficients that well approximate the solution to \eqref{eq:self-expression-optimization}, which justifies its ability in obtaining the desired subspace-preserving and denser 
connection properties.\footnote{An analysis of its approximation power is left as future work.}

An important benefit of the design 
in \eqref{eq:SENet-formulation} is that the computation of the self-expressive coefficient matrix for a given data matrix $X$ can be made very efficient. In particular, instead of evaluating $f(\x_i, \x_j; \Theta)$ for all possible (\ie, $N^2$ number of) pairs of $(\x_j, \x_i)$, one may evaluate $\u(\cdot, \Theta_u)$ and $\v(\cdot, \Theta_v)$ separately for all columns of $X$, which can be parallelized.
After that, the coefficient matrix can be obtained by computing inner product between pairs of $(\u_j, \v_i)$, which again can be parallelized and computed very efficiently. Such a property also allows us to train the network efficiently as we explain in the next subsection.

\myparagraph{Comparison to Self-attention Models}
The network architecture in \eqref{eq:SENet-formulation} bears a close resemblance to the \emph{self-attention} model in Transformer \cite{Vaswani:NIPS17}, Non-Local Neural Networks~\cite{Wang:CVPR18}, and Graph Attention Networks~\cite{Velick:ICLR18}, etc., which also aim to compute self-expressive coefficients for a set of signal (\eg, sequence, image, video, nodes on graph) representations.
However, we note that our choice of architecture in \eqref{eq:SENet-formulation} has several favorable properties over the self-attention models.
\begin{itemize}[leftmargin=*,topsep=0.25em,noitemsep]
    \item The functions $\u(\cdot)$ and $\v(\cdot)$ in self-attention models are linear maps, while we use MLPs for our SENet.
    This design is to increase the expressive power of SENet to gain universal approximation ability so that it can easily approach the optimal solution for the convex formulation in \eqref{eq:self-expression-optimization} and hence enjoy subspace-preserving property.

    \item The self-attention model usually adopts a normalization factor such that each self-expression is given by a \emph{convex} combination. Such a requirement is, however, too restrictive for our purpose: for sample points that lie in a vertex of the convex hull of sample points in one of the subspaces, they cannot be expressed as a convex combination of other points.
    In such cases, self-attention models cannot produce subspace-preserving solutions.

    \item We adopt a soft-thresholding operator at the output of SENet, borrowed from learned sparse optimization networks such as LISTA~\cite{Gregor:ICML10} and ISTA-Net \cite{Zhang:CVPR18}, to enforce sparsity of the output.
    This is to account for the fact that the solution to the model in \eqref{eq:self-expression-functional-optimization} with the elastic net regularization in \eqref{eq:elastic-net-regularization} is expected to be sparse (due to the $\ell_1$ norm inside). 
\end{itemize}


\subsection{Training}
\label{subsec:training}

We train 
SENet in \eqref{eq:SENet-formulation} via solving the following optimization problem:
\begin{align}
\label{eq:def-L}
\min_\Theta \L(X; \Theta) := \sum_{j=1}^N \ell(\x_j, X; \Theta),
\end{align}
where $\ell(\x_j, X; \Theta)$ is the objective function in \eqref{eq:self-expression-functional-optimization}, i.e.,
\begin{align}
\begin{split}
\ell(\x_j, X; \Theta) :&= \frac{\gamma}{2}\| \x_j - \sum_{i\ne j} f(\x_i, \x_j; \Theta) \x_i \|_2^2 \\
                       &+\sum_{i \ne j} r\big(f(\x_i, \x_j; \Theta)\big).
\label{eq:self-expression-objective}
\end{split}
\end{align}
Then, the network parameters $\Theta$ can be learned by Stochastic Gradient Descent (SGD).
We summarize the algorithm (assuming that batch size is $1$ for simplicity) in Algorithm~\ref{alg:naive}.

Since the loss $\ell(\x_j, X; \Theta)$ depends on the entire data $X$ (for any fixed $\x_j$), the memory requirement for Algorithm~\ref{alg:naive} scales linearly with the number of data points. This restricts the ability of the algorithm to handle very large scale data. Next, we present a two-pass algorithm that is equivalent to Algorithm~\ref{alg:naive} but with constant memory complexity. 

\begin{algorithm}[t]
	\caption{A Naive SGD Algorithm for Training SENet}
	\label{alg:naive}
	\begin{algorithmic}[1]
		\STATE \textbf{Input:} Dataset $X \in \mathbb{R}^{D \times N}$,  model parameters $\gamma > 0$ and $\lambda \in [0, 1]$, number of iterations $T$, learning rate $\eta$
		\STATE \textbf{Initialization:} Random initialize SENet parameters $\Theta$
			\FOR {each $t \in \{1, \cdots, T\}$}
			    \STATE \label{step:batch-sample}Sample a data point $\x_j$ from $X$
			    \STATE \emph{\# Forward propagation to compute the loss}
			    \STATE Compute $\u_j \doteq \u(\x_j, \Theta_u)$ 
			    \STATE Load data $X$ and compute $V \doteq [\v_1, \ldots, \v_N]$, where $\v_i \doteq \v(\x_i, \Theta_v)$
			    \STATE Compute $f(X, \x_j; \Theta) \doteq \alpha \mathcal{T}_b(V^\top \u_j)$ 
    			\STATE Compute $\ell(\x_j, X; \Theta)$ from $f(X, \x_j; \Theta)$ by~\eqref{eq:self-expression-objective}
    			\STATE \emph{\# Backward propagation to compute the gradient}
    			\STATE \label{step:naive-gradient}Compute $d\Theta \doteq \frac{\partial \ell(\x_j, X;  \Theta)}{\partial \Theta}$
    		    \STATE \emph{\# Gradient descent to update $\Theta$}   	
    			\STATE Set $\Theta \leftarrow \Theta - \eta \cdot d\Theta$
			\ENDFOR
		\STATE \textbf{Output:} SENet with trained weights.
		\end{algorithmic}
\end{algorithm}

\myparagraph{Two-pass SGD Algorithm}
To derive our algorithm, we compute the gradient in step~\ref{step:naive-gradient} of Algorithm~\ref{alg:naive} as
\begin{multline}
\label{eq:derivative}
\frac{\partial \ell(\x_j, X; \Theta)}{\partial \Theta}
= \\\sum_{i\ne j} \Big( r'\big(f(\x_i,\x_j; \Theta)\big) - \langle \x_i, \q_j \rangle \Big) \frac{\partial f(\x_i,\x_j; \Theta)}{\partial \Theta},
\end{multline}
where
\begin{equation}\label{eq:def_q}
    \q_j := \gamma \Big(\x_j - \sum_{i\ne j} f(\x_i,\x_j; \Theta)\x_i\Big),
\end{equation}
and $r'(\cdot)$ denotes the derivative\footnote{As $r(t)$ is not differentiable at $t = 0$, we set $r'(0) = 0$ which is in the sub-differential of $r(t)$ at $t=0$.} of $r(\cdot)$.
Observe that if the vector $\q_j$ in \eqref{eq:derivative} is given, then the right hand side of \eqref{eq:derivative} is a weighted sum of gradient computed at each data point $\x_i$ for $i = 1, \cdots, N$. Therefore, it can be accumulated in an online fashion with constant space requirement (see step~\ref{step:two-pass-second} - \ref{step:two-pass-second-end}, Algorithm~\ref{alg:two-pass}).
Moreover, although $\q_j$ is unknown, it can be computed by performing a separate forward propagation (and no backward propagation is needed).
In particular, $\q_j$ can be computed by subtracting the summation $\sum_{i\ne j} f(\x_i,\x_j; \Theta)\x_i$ from $\x_j$, where the summation term can be accumulated in an online fashion with constant space requirement as well (see step~\ref{step:two-pass-first} - \ref{step:two-pass-first-end}, Algorithm~\ref{alg:two-pass}).
Overall, this leads to a two-pass algorithm for training SENet as described in Algorithm~\ref{alg:two-pass}.

Since the memory requirement for Algorithm~\ref{alg:two-pass} does not scale with the number of data points, in principle it can handle arbitrarily large datasets.

\begin{algorithm}
	\caption{A Two-pass Algorithm for Training SENet}
	\label{alg:two-pass}
	\begin{algorithmic}[1]
		\STATE \textbf{Input:} Dataset $X \in \mathbb{R}^{D \times N}$,  model parameters $\gamma > 0$ and $\lambda \in [0, 1]$, number of iterations $T$, learning rate $\eta$
		\STATE \textbf{Initialization:} Random initialize SENet parameters $\Theta$\!\!\!
		\FOR {each $t \in \{1, \cdots, T\}$}
			\STATE \label{step:two-pass-batch-sample}Sample a data point $\x_j$ from $X$
			\STATE Compute $\u_j \doteq \u(\x_j, \Theta_u)$ 
		    \STATE \label{step:two-pass-first}\emph{\# First pass (forward only): compute $\q_j$}
			\STATE Initialize $\bar\x = \0$
			\FOR {each $i \in \{1, \cdots, j-1, j+1, \cdots, N\}$}\label{step:two-pass-first-iterate}
				\STATE Load data $\x_i$ and compute  $\v_i \doteq \v(\x_i, \Theta_v)$ 
			    \STATE Compute $f(\x_i,\x_j; \Theta) = \alpha \mathcal{T}_b(\u_j^\top \v_i)$ 
				\STATE Set $\bar\x \leftarrow \bar\x + f(\x_i,\x_j; \Theta)\x_i$
			\ENDFOR
			\STATE Set $\q_j = \gamma(\x_j - \bar{\x})$ \label{step:two-pass-first-end}
 		    \STATE \label{step:two-pass-second}\emph{\# Second pass: compute gradient $d\Theta$}   	
 		    \STATE Initialize $d\Theta = \0$
			\FOR {each $i \in \{1, \cdots, j-1, j+1, \cdots, N\}$}\label{step:two-pass-second-iterate}
			    \STATE Load data $\x_i$ and compute  $\v_i \doteq \v(\x_i, \Theta_v)$ 
			    \STATE Compute $f(\x_i,\x_j; \Theta) = \alpha \mathcal{T}_b(\u_j^\top \v_i)$ 
				\STATE Set $d\Theta \leftarrow d\Theta + \Big( r'(f(\x_i,\x_j; \Theta)) - \langle \x_i, \q_j \rangle \Big) \frac{\partial f(\x_i,\x_j; \Theta)}{\partial \Theta}$
			\ENDFOR \label{step:two-pass-second-end}
		    \STATE \emph{\# Gradient descent to update $\Theta$}   	
			\STATE Set $\Theta \leftarrow \Theta - \eta \cdot d\Theta$
		\ENDFOR
		\STATE \textbf{Output:} SENet with trained weights.
		\end{algorithmic}
		
\end{algorithm}
\vspace{-2pt}

\section{Experiments}
\label{sec:experiments}

We conduct extensive experiments on both synthetic data and real world benchmark datasets to evaluate the performance of SENet.

\myparagraph{Network Architecture} For both the query and key 
networks in \eqref{eq:SENet-formulation-u} and \eqref{eq:SENet-formulation-v}, we use a three-layer MLP with ReLU and $\text{tanh}(\cdot)$ as the activation functions for hidden layers and the output layer, respectively.
The number of hidden units in each layer of the MLPs are $\{1024, 1024, 1024\}$, and the output dimension $p$ is $1024$. By using $\text{tanh}(\cdot)$ as the output layer activation, the inner product of the output vectors $\u_j$ and $\v_i$ is bounded by $p$, \ie, $\u_j^\top \v_i \in (-p, p)$.
Therefore, we use a small scalar multiplier $\alpha=\frac{1}{p}$ 
as in \eqref{eq:SENet-formulation} to scale down the output of SENet. We use the Adam~\cite{Kingma:ICLR2014} optimizer with an initial learning rate of $10^{-3}$ and use the cosine annealing learning rate decay~\cite{Loshchilov:ICLR17} with gradient clipping.

\myparagraph{Metrics} Given a self-expressive coefficient matrix $C$, we use the subspace recovery error (SRE), defined as the proportion of the $\ell_1$ norm of $C$ that comes from the wrong subspace, to measure the subspace-preserving property of $C$.
In addition, we use the algebraic connectivity (CONN) \cite{Mohar:GTCA91}, defined as the second smallest eigenvalue of the normalized graph Laplacian of each ground-truth class minimized over all classes, to measure the connectedness of the affinity graph.
As discussed in Subsection~\ref{subsec:model}, we desire that $C$ has low SRE and high CONN.
We refer the reader to \cite{You:CVPR16-SSCOMP} for a detailed explanation of these two quantities.

To evaluate the clustering performance, we report clustering accuracy (ACC), normalized mutual information (NMI) and adjusted rand index (ARI) which are commonly used in the literature (see e.g., \cite{Yu:NIPS20} for a definition).


\subsection{Experiments on Synthetic Data}
\label{sec:Synthetic Experiments}

\begin{figure}[htbp]
\small
\vspace{-5pt}
\centering
	\subfigure[{\small $\small C^{(100)}_{tr}$ (64\%)}]{\includegraphics[clip=true,trim=8 8 0 5,width=0.315\columnwidth]{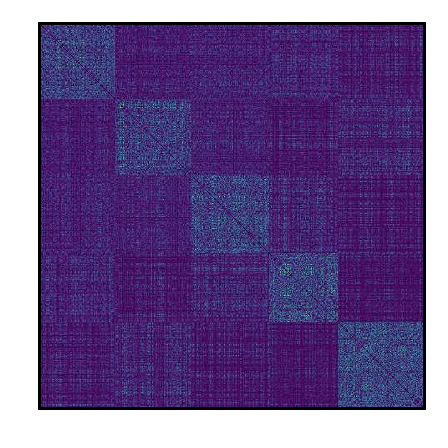}}
	\subfigure[{\small $\small C^{(300)}_{tr}$ (23\%)}]{\includegraphics[clip=true,trim=8 8 0 5, width=0.315\columnwidth]{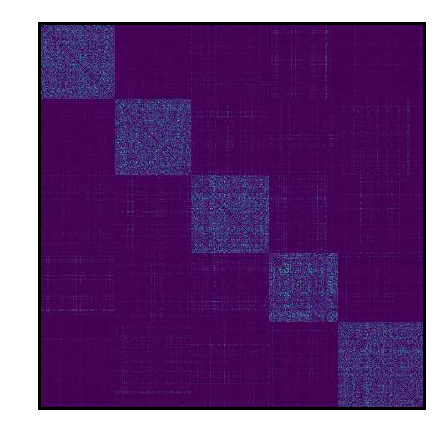}}
	\subfigure[{\small $\small C^{(500)}_{tr}$ (10\%)}]{\includegraphics[clip=true,trim=8 8 0 5, width=0.315\columnwidth]{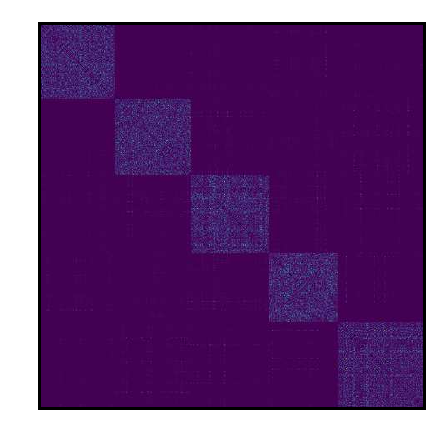}} \\
	\subfigure[{\small $\small C^{(100)}_{ts}$ (66\%)}]{\includegraphics[clip=true,trim=8 8 0 5, width=0.315\columnwidth]{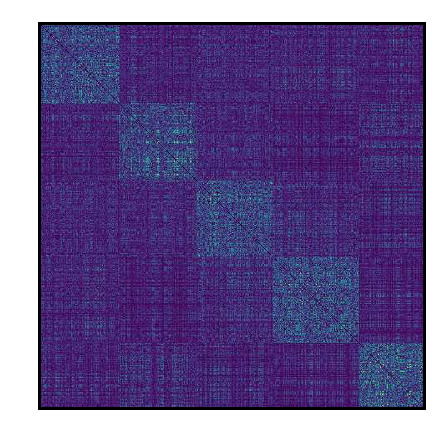}}
	\subfigure[{\small $\small C^{(300)}_{ts} $ (37\%)}]{\includegraphics[clip=true,trim=8 8 0 5, width=0.315\columnwidth]{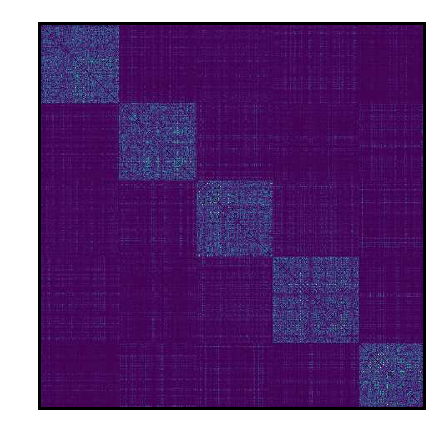}}
	%
	\subfigure[{\small $\small C^{(500)}_{ts}$ (24\%)}]{\includegraphics[clip=true,trim=8 8 0 5, width=0.315\columnwidth]{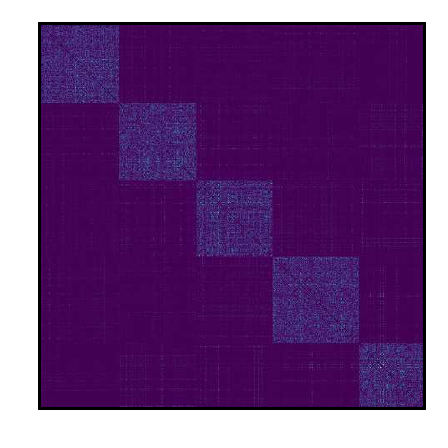}}
	\caption{Visualization of self-expressive coefficients computed by SENet trained with $\{100, 300, 500\}$ iterations on synthetic data where the percentage number in bracket is SRE.} 
	\label{fig:data-visual-synthetic-vs-iter}
\end{figure}

\myparagraph{Visualization of Self-expressive Coefficients} 
We demonstrate the ability of SENet to produce self-expressive coefficients and generalize to out-of-sample data on synthetic data. For that purpose, we generate a synthetic dataset as in \cite{You:CVPR16-SSCOMP}, 
where $5$ subspaces of dimension $6$ are sampled uniformly at random in the ambient space $\RR^{15}$ (\ie, $n=5$, $d=6$ and $D=15$),
and $200$ points are sampled uniformly at random on the unit 
sphere of each subspace. We randomly select $500$ data points as training data $X_{tr}$ and the remaining $500$ data points as testing data $X_{ts}$. We set the parameters $\gamma=50.0$ and $\lambda=0.9$ and use Algorithm \ref{alg:naive} to train our SENet on $X_{tr}$ with maximum iteration $T_{max}=500$. Then we take the trained SENet at the $t$-th iteration to evaluate and infer the matrices of self-expressive coefficients $C^{(t)}_{tr}$ and $C^{(t)}_{ts}$ on $X_{tr}$ and $X_{ts}$, respectively. 
A visualization of $|C^{(t)}_{tr}|$ and $|C^{(t)}_{ts}|$ is given as colored images in Fig.~\ref{fig:data-visual-synthetic-vs-iter}. We  observe of that SENet is able to efficiently learn self-expressive coefficients that are approximately subspace-preserving after only a few hundred iterations and that the trained SENet is able to infer self-expressive coefficients for out-of-sample data with reasonably good quality. Note that spectral clustering could yield perfect result after training with 300 iterations.


\myparagraph{Comparing SENet to EnSC} 
We demonstrate the ability of SENet to approximate the solution to \eqref{eq:self-expression-optimization} with $r(\cdot)$ being the elastic net regularization function in \eqref{eq:elastic-net-regularization}, which is a method known as EnSC \cite{You:CVPR16-EnSC}. For that purpose, we use the same parameters $\gamma=50.0$ and $\lambda=0.9$ for SENet and EnSC models, so that they solve the same optimization problems except that EnSC directly optimizes over the self-expressive coefficients while SENet optimizes over the parameters of a network that generates the coefficients.

We sample $5$ subspaces of dimension $6$ in the ambient space $\RR^9$ (\ie, $n=5$, $d=6$ and $D=9$), then sample $N_i$ data points from the unit sphere of each subspace with $N_i \in \{20, 100, 200, 1000, 2000\}$. 
We measure the difference between EnSC and SENet solutions by reporting the total loss ($\L$) in \eqref{eq:def-L}, as well as the reconstruction loss and regularization loss:
\begin{align}
\L_{rec}&\doteq\sum_{j=1}^N \| \x_j - \sum_{i\ne j} f(\x_i, \x_j; \Theta) \x_i \|_2^2,\\
\L_{reg}&\doteq \sum_{j=1}^N \sum_{i \ne j} r(f(\x_i,\x_j; \Theta)).
\end{align}
We also report SRE, CONN and ACC. 
The results are shown in Table \ref{table:equivalence-test}. We can see that the difference between the solution by SENet and EnSC is relatively small, indicating the strong approximation power of the SENet architecture. On the other hand, such a difference increases with $N_i$, showing that a larger (e.g., deeper and wider) network may be needed. By examining the values of SRE and CONN we can see that such a difference causes higher subspace-preserving error, but it helps improve the connectivity of the affinity graph.

%
To evaluate the generalization ability of the trained SENet, we prepare a set of test data that consists of $N_i$ data points per subspace sampled uniformly at random from the union of subspaces model that is used to generate the training data.
%
%
Then, the trained SENet is used to directly infer the self-expressive coefficients on test data. 
The results are reported in the rows ``SENet test'' of  Table~\ref{table:equivalence-test}.
We can see that the trained SENet shows increasingly better ability to detect subspace structures when the number of data points per subspace is increased.
Moreover, we can see that while $\L_{reg}$ is similar in scale to that given by SENet on the train data, the $\L_{rec}$ is significantly higher.
This shows that the generalization ability of SENet is on detecting subspace structures, not on the reconstruction.

\renewcommand{\arraystretch}{1.1}
\begin{table}[!htbp]
	\centering
\tiny
	\begin{tabular}{l|l|c c c c c c}
		\hline
		\multirow{2}{*}{$N_i$} &\multirow{2}{*}{Methods}&\multicolumn{6}{|c }{Metrics}\\
		&\multirow{2}{*}{} & $\L$ & $\L_{rec}$ &$\L_{reg}$ & ACC (\%) &SRE (\%) &CONN\\
		\hline
		\multirow{3}{*}{20}
		&EnSC   &135.127&0.107&132.442& 72.0&49.611&0.178\\
		&SENet  &135.132&0.109&132.416& 71.0&49.720&0.178\\
		&SENet test &1830.107 &72.007 &29.937 &65.0 &58.384 &0.318 \\
		\hline
		\multirow{3}{*}{100}
		&EnSC   &559.943&0.526&558.009&93.0&27.370&0.163\\
		&SENet  &559.972&0.531&558.022&92.8&27.501&0.165\\
		&SENet test & 2935.424 & 89.325 & 702.309  &  79.0  & 56.897  & 0.387 \\
		\hline
		\multirow{2}{*}{200}
		&EnSC   &1053.086&0.526&1040.097&96.6&20.067&0.155\\
		&SENet  &1053.369&0.531&1040.097&96.0&20.195&0.159\\
		&SENet test  &17826.273&599.099&2848.779&84.1&56.256&0.398\\
		\hline
		\multirow{2}{*}{1000}
		&EnSC   &4884.876&2.095&4832.508&99.4&6.493&0.126\\
		&SENet  &4932.907&2.205&4877.781&99.5&9.132&0.155\\
		&SENet test  &30037.012&887.323 &7853.945&92.3&36.054&0.236\\
		\hline
		\multirow{2}{*}{2000}
		&EnSC   &9576.154&3.958&9477.197&99.7&4.580&0.108\\
		&SENet  &10025.874&4.592&9911.074&99.4&13.555&0.201\\
		&SENet test  &44458.734&1453.790 &8113.975&97.4&21.863&0.220\\
		\hline
	\end{tabular} 	
	\\[4pt]
	\caption{{Comparing SENet to EnSC on synthetic data}}
	\label{table:equivalence-test}
	\vspace{-10pt}
\end{table}

\subsection{Experiments on Real World Datasets}
\label{sec:Real Experiments}

We further evaluate the performance of SENet on four larger 
benchmark datasets: 
MNIST~\cite{LeCun:PIEEE1998}, Fashion-MNIST \cite{Xiao:FashionMNIST19}, CIFAR-10 \cite{Krizhevsky:CIFAR2009} and Extended MNIST (EMNIST) \cite{Cohen:arXiv17}.\!\!


%
%
\textbf{MNIST} contains 70,000 grey-scale images of handwritten digits ``0'' to ``9'', which we denote as MNIST-full. The MNIST-full is divided into MNIST-train and MNIST-test, consisting of 60,000 and 10,000 images, respectively.
%
%
%
\textbf{Fashion-MNIST} contains 70,000 grey-scale images of various types of fashion products, denoted as Fashion-MNIST-full.
Fashion products (\eg, coat, trouser, shirt, dress, bag, etc.) with different styles correspond to 10 categories.
Similar to MNIST, Fashion-MNIST-full is divided into Fashion-MNIST-train and Fashion-MNIST-test, consisting of 60,000 and 10,000 images, respectively.
%
%
%
\textbf{EMNIST} contains grey-scale images of handwritten digits and letters where $190,\!998$ images of the 26 lower case letters are used for the clustering problem with $26$ categories. 
For these three datasets, we compute a feature vector of dimension 3,472 using the scattering convolution network \cite{Bruna:PAMI13}, which extracts translational invariant and deformation stable features, and then reduce the dimension to $500$ using PCA.
\textbf{CIFAR-10} contains 60,000 color images in 10 classes, where each image is of size $32\times 32$.
For CIFAR-10, we use the feature representation extracted by MCR$^2$ \cite{Yu:NIPS20}, which learns a union-of-subspace representation from data with self-supervised learning.
All feature vectors are normalized to have unit $\ell_2$ norm.\footnote{For EMNIST, we also remove the mean after PCA as in \cite{You:ECCV18}.} 

To produce a segmentation from the self-expressive coefficient matrix, we compute an affinity matrix by either a) constructing a 3-nearest neighbor graph from the columns of $C$ as in \cite{You:ECCV18} (for MNIST, Fashion-MNIST and EMNIST),
or b) using $|C| + |C^\top|$ (for CIFAR-10). Then, spectral clustering is applied to the affinity matrix.\footnote{For MNIST and FashionMNIST, we use the eigenvectors corresponding to the 15 smallest eigenvalues of graph Laplacian to perform $k$-means.}

\begin{figure*}[htbp]
\small
\centering
\subfigure[MNIST]{
\centering
\includegraphics[clip=true,trim=2 0 5 5, width=0.465\columnwidth]{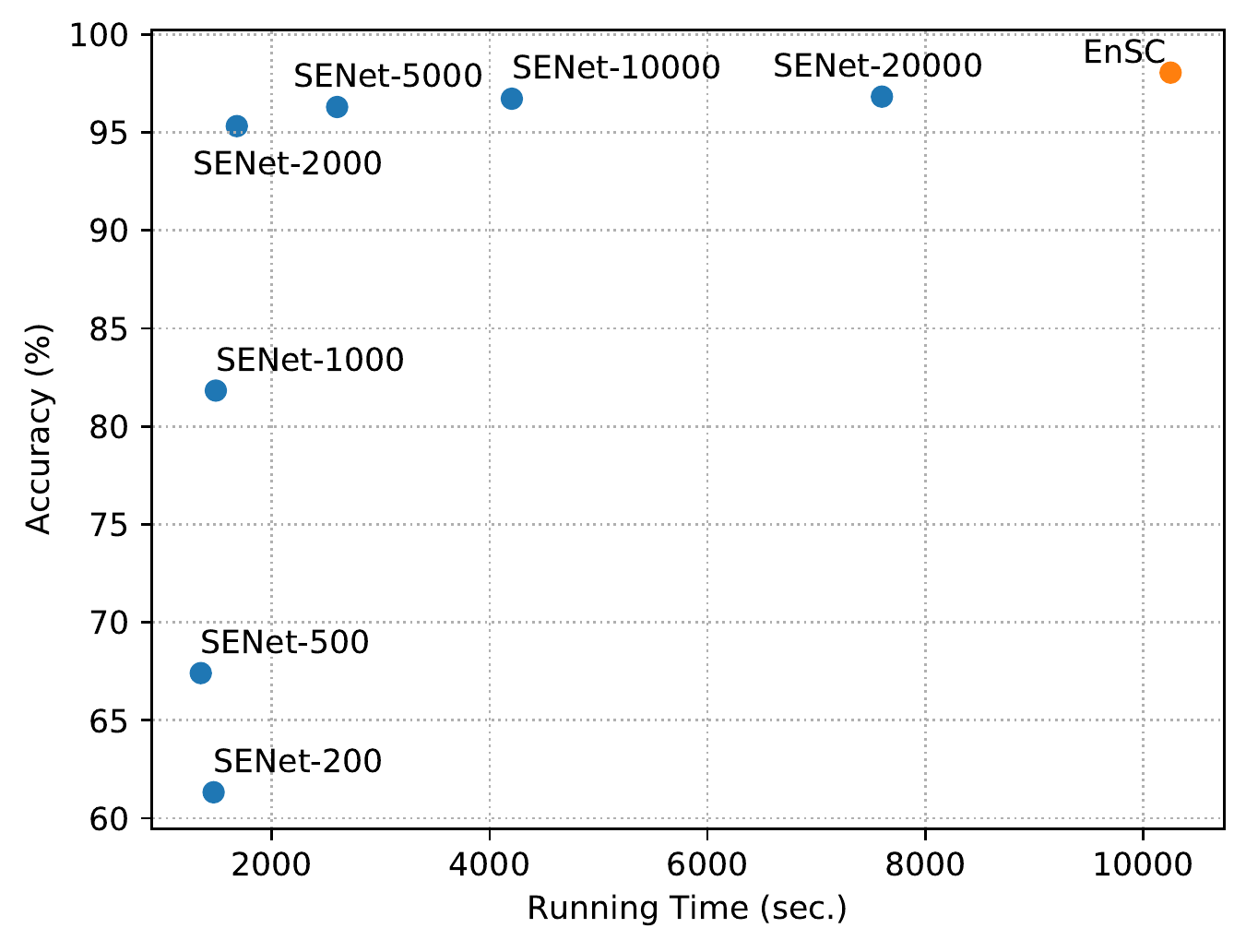}}
\subfigure[Fashion-MNIST]{
\centering
\includegraphics[clip=true,trim=2 0 5 5, width=0.465\columnwidth]{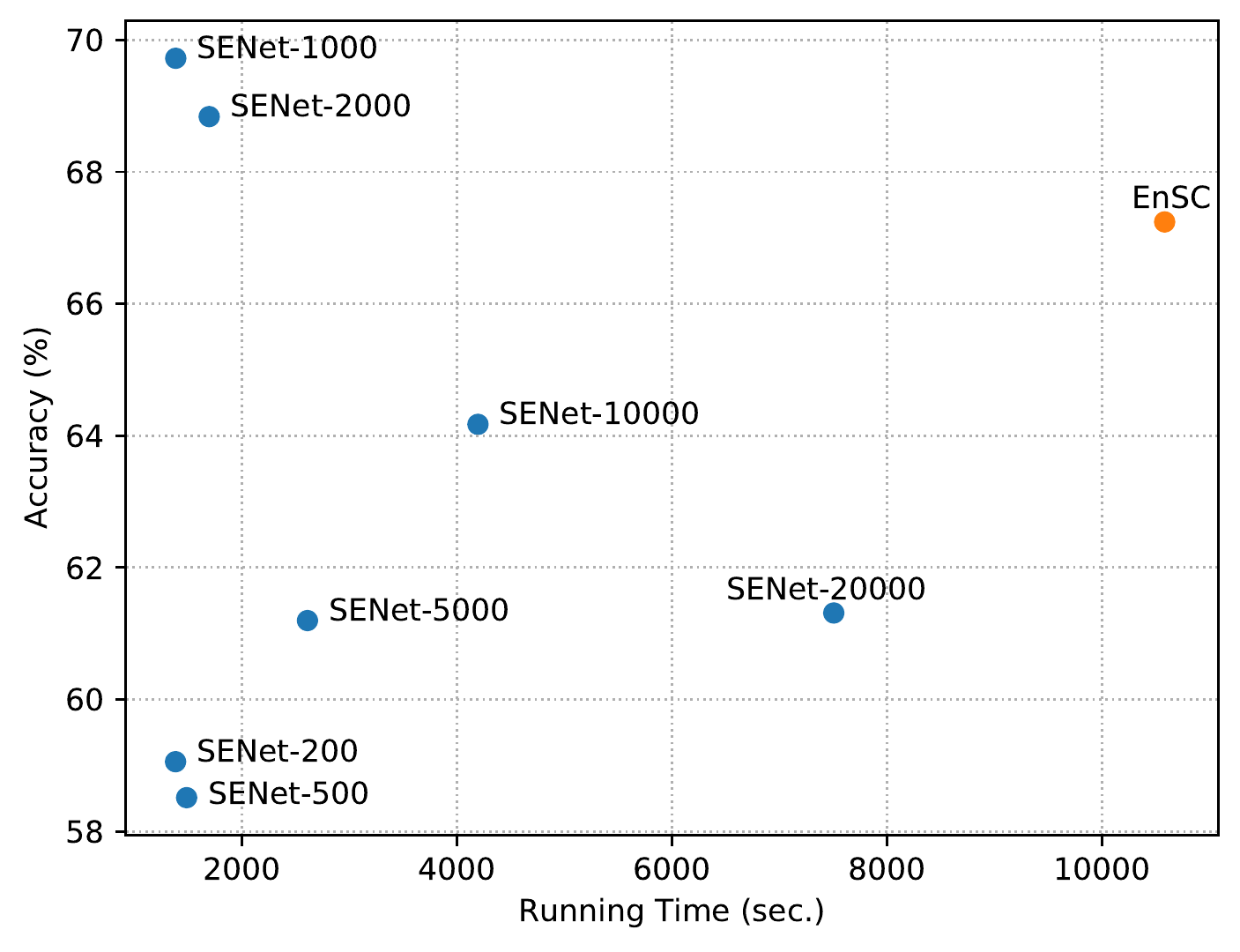}}
\subfigure[CIFAR-10]{
\centering
\includegraphics[clip=true,trim=2 0 5 5, width=0.465\columnwidth]{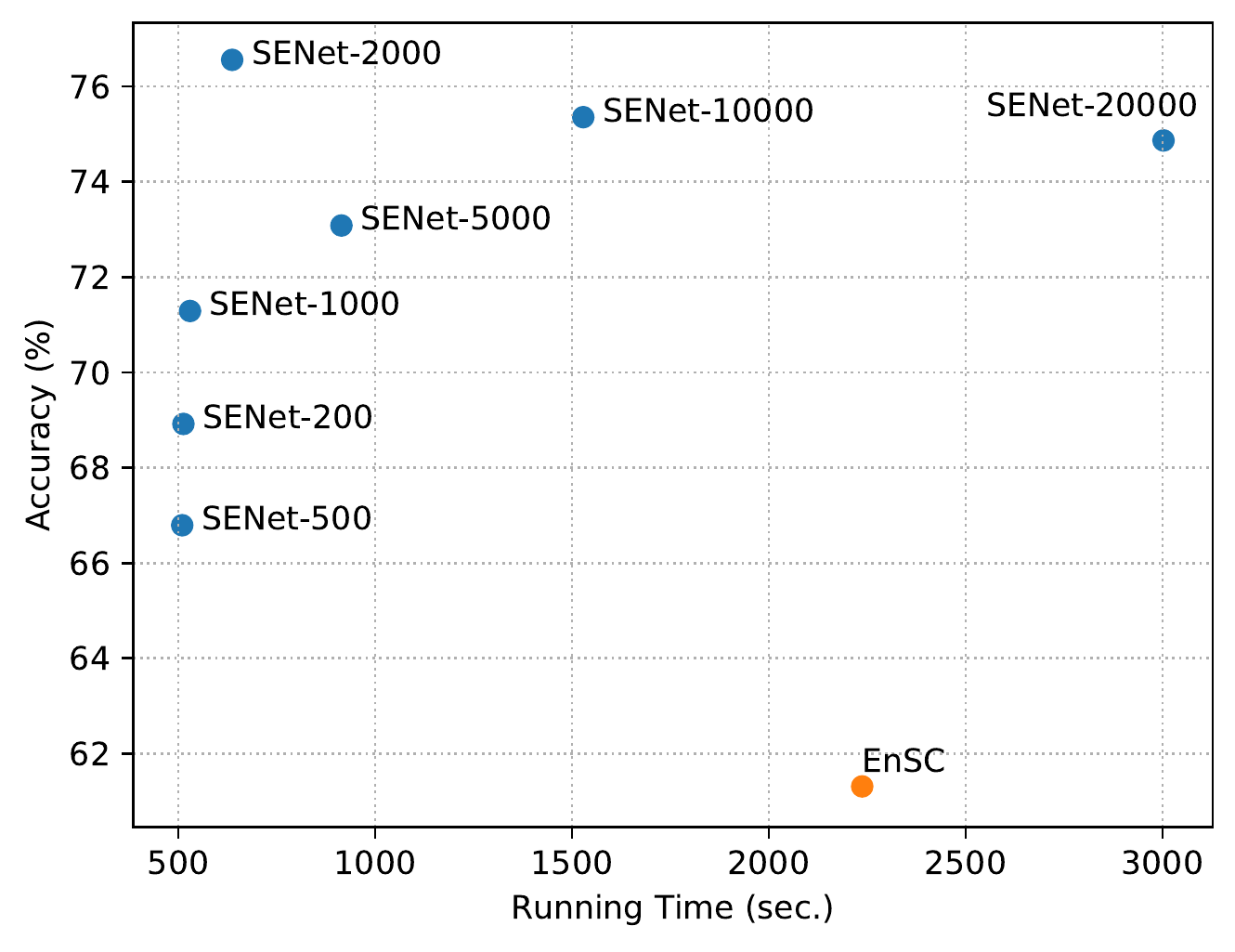}	}
\subfigure[EMNIST]{
\centering
\includegraphics[clip=true,trim=2 0 5 5, width=0.465\columnwidth]{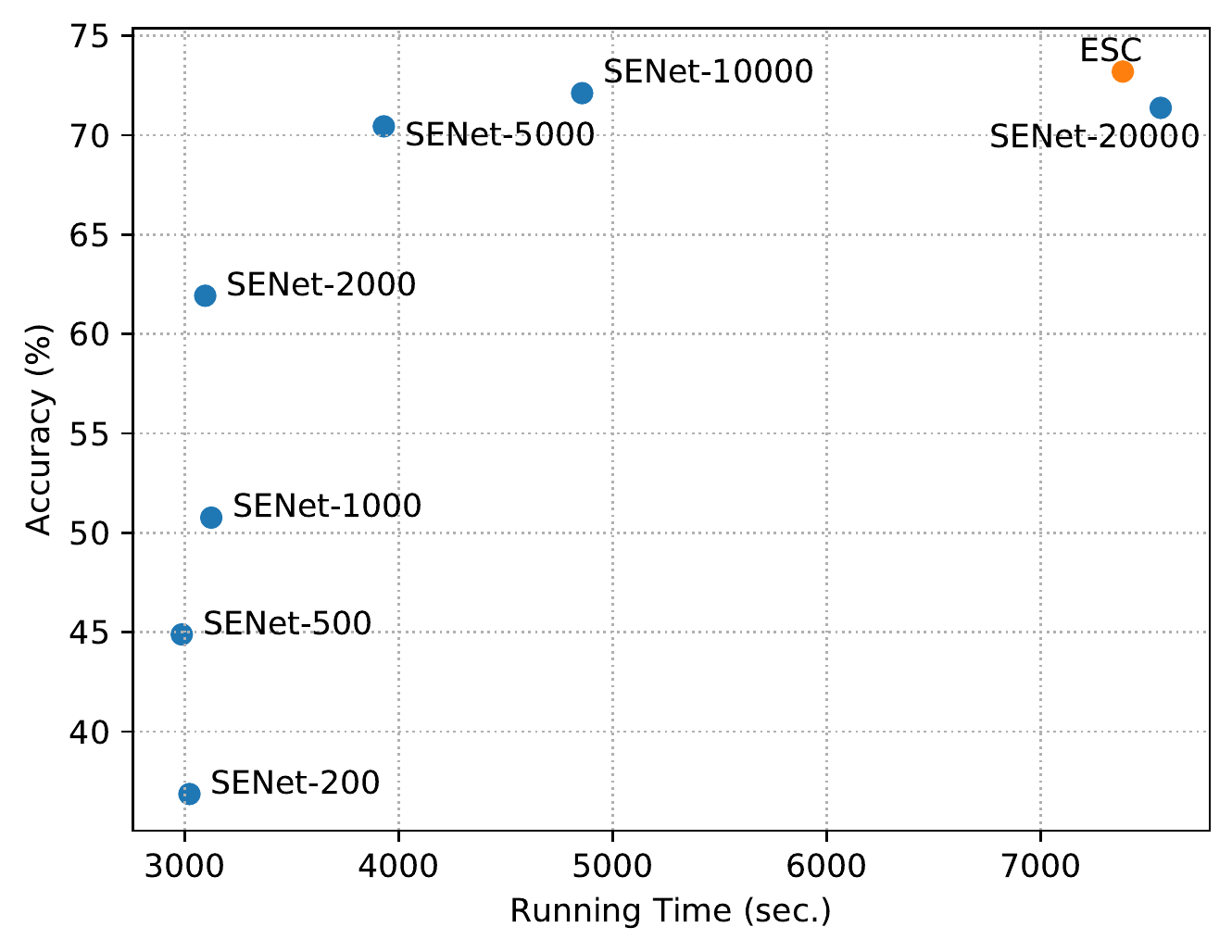}	}
\caption{Clustering accuracy vs. training time with varying training data sizes. SENet-$N$ denotes SENet trained on $N$ data points.
}
\label{fig:Real-Dataset-Diff-DictSize}
\end{figure*}

\myparagraph{Generalization Performance of SENet}
%
We evaluate the generalization ability of SENet to out-of-sample data using MNIST and Fashion-MNIST.
Specifically, we select $N \in \{200, 500, 1000, 2000, 5000, 10000, 20000\}$ data points uniformly at random from MNIST-train and train SENet for 100,000 iterations with a batch size fixed to 100 (likewise for Fashion-MNIST-train).
Then, we take MNIST-test as test data for which the trained SENet is used to generate self-expressive coefficients and apply spectral clustering on the induced affinity for producing a segmentation (likewise for Fashion-MNIST-test).
For EnSC, we directly compute the self-expressive coefficients on MNIST-test and Fashion-MNIST-test.

Experimental results are reported in Table~\ref{tab:test-out-of-sample-MNIST-FashionMNIST}. We can see that with the increasing amount of training data, SENet is able to approach or surpass the performance of EnSC, which is directly optimized on the test data. This confirms that the trained SENet enjoys a promising generalization ability to out-of-sample data in real world datasets.

\renewcommand{\arraystretch}{1.1}
\begin{table}[!htbp]
	\centering
	\small
    \resizebox{0.46\textwidth}{!} {
	\begin{tabular}{l|l|c c|c c}
		\hline
		\multirow{2}{*}{Methods} & \multirow{2}{*}{Train Data: \# } &\multicolumn{2}{|c|}{MNIST-test}          &\multicolumn{2}{|c}{Fashion-MNIST-test} \\
		\multirow{2}{*}{} & & ACC (\%) & SRE (\%) & ACC (\%) & SRE (\%) \\
		\hline
		EnSC    & NA    &\bf{97.15} &4.455                &60.55 &21.712 \\
        \hline
		\multirow{7}{*}{SENet}
		& 200              &77.22 &14.260                    &55.41 &26.299\\
		& 500              &82.60 &8.846                     &63.65 &24.430\\
		& 1000             &80.87 &7.290                     &\bf{70.46} &23.502\\
		& 2000             &95.45 &5.131                	 &58.71 &22.197\\
		& 5000             &95.80 &4.785                     &60.67 &21.109\\
		& 10000            &\underline{96.66} &\underline{4.121}          &62.92 &\bf{20.385}\\
        & 20000            &96.25 &\bf{3.978}          &\underline{64.64} &\underline{20.442} \\
		\hline
	\end{tabular} 	
	}
	\\[4pt]
	\caption{Generalization performance of SENet on MNIST-test and Fashion-MNIST-test. 
	}
	\label{tab:test-out-of-sample-MNIST-FashionMNIST}
\end{table}

\begin{table*}[!htbp]
	\centering
	\footnotesize
    \resizebox{0.78\textwidth}{!} {
	\begin{tabular}{l|c c c|c c c|c c c|c c c}
		\hline
		\multirow{2}{*}{Methods}  &\multicolumn{3}{|c|}{MNIST-full}          &\multicolumn{3}{c|}{Fashion-MNIST-full}
		&\multicolumn{3}{c|}{CIFAR-10}
		&\multicolumn{3}{c}{EMNIST}\\
		& ACC & NMI & ARI
		& ACC & NMI & ARI
		& ACC & NMI & ARI
		& ACC & NMI & ARI\\
		\hline
		$k$-means \cite{MacQueen-1967}	&0.541	&0.507	&0.367	&0.505	&0.578	&0.403	&0.525	&0.589	&0.276
		&0.459 &0.438 &0.316 \\
		\hline
		Spectral \cite{Shi-Malik:PAMI00} 	&0.728	&0.856	&0.667	&0.625	&{0.700}  &0.494	&0.455	&0.574	&0.256 &0.662 &0.769 &0.654 \\
		\hline
		JULE \cite{Yang:CVPR16} 	&0.964	&0.913	&0.927	&0.563	&0.608	&-	&0.272	&0.192	&0.138 &- &- &-\\
		\hline
		DEC \cite{Xie:ICML16}	    &0.863	&0.834	&-	&0.518	&0.546	&-	&0.301	&0.257	&0.161 &- &- &- \\
		\hline
		DAC \cite{Chang:ICCV17-DAC}	&\underline{0.978} 	&0.935	&\underline{0.949}	&-	&-	&-	&0.522	&0.396	&0.306 &- &- &- \\
		\hline
		DEPICT \cite{Ghasedi:ICCV17-DEPICT}	&0.965	&0.917	&-	&0.392	&0.392	&-	&-	&-	&- &- &- &- \\
		\hline
		ClusterGAN \cite{Mukherjee:AAAI19-ClusterGAN}	&0.905	&0.890	&-	&0.662	&0.645	&-	&-	&-	&- &- &- &-\\
		\hline
		DSCDAN \cite{Yang:CVPR19}	&\underline{0.978}	&\underline{0.941}	&-	&0.662	&0.645	&-	&-	&-	&- &- &- &-\\
		\hline
		DCCM \cite{Wu:ICCV19}	&-	&-	&-	&-	&-	&-	&0.623	&0.496	&0.408 &- &- &-\\
		\hline
		SSC-OMP \cite{You:CVPR16-SSCOMP}	&0.928	&0.842	&0.849	&0.274	&0.421	&0.196	&0.326	&0.498 &0.196 &0.654 &0.661 &0.634\\
		\hline
		NCSC \cite{Zhang:ICML19}	&0.941	&0.861	&0.875	&\bf{0.721}	&0.686	&\bf{0.592}	&-	&-	&- &- &- &-\\
		\hline
		EnSC \cite{You:CVPR16-EnSC}	&\bf{0.980}	&\bf{0.945}	&\bf{0.957}	&0.672	&\underline {0.705}	&\underline{0.565}	&0.613	&0.601	&0.430 &T &T &T\\
		\hline
		ESC~\cite{You:ECCV18} &0.971 &0.925 &0.936 &0.668 &\bf{0.708} &0.556	&\underline{0.653} &\underline{0.629} &\underline{0.438} &\bf{0.732} &\bf{0.825} &\underline{0.759}\\
		\hline
		SENet	&0.968	&0.918	&0.931	&\underline{0.697}	&0.663	&0.556	&\bf{0.765}	&\bf{0.655}	&{\bf 0.573} &\underline{0.721} &\underline{0.798} &\bf{0.766}\\
		\hline
	\end{tabular} 	
    }
	\\[4pt]
	\caption{Image clustering results on MNIST-full, Fashion-MNIST-full, CIFAR-10 and EMNIST. The best results are in bold font and the second best results are underlined. `T' means the computation time exceeds 24 hours.}
	\label{tab:results-on-three datasets compare to SOTA}
\end{table*}

\myparagraph{Subspace Clustering on Large-Scale Datasets}
%
We demonstrate that SENet can effectively handle large-scale datasets MNIST-full (70k), Fashion-MNIST-full (70k), CIFAR-10 (60k) and EMNIST (190k).
For each dataset, we randomly select $N$ points to train SENet, then apply the trained SENet to generate self-expressive coefficients on the entire dataset. At the end, spectral clustering is applied to obtain the segmentation.
In EnSC and SENet, we use $\gamma=200.0$ and $\lambda=0.9$ for MNIST, Fashion-MNIST and CIFAR-10, and $\gamma=150.0$ and $\lambda=1.0$ for EMNIST.

In Fig.~\ref{fig:Real-Dataset-Diff-DictSize}, we report the training time and clustering accuracy with varying $N$.
The experiments are conducted on a single NVIDIA GeForce 2080Ti GPU (for EMNIST) or 1080Ti GPU (for all other datasets).
We also compare with EnSC, for which the active support solver in \cite{You:CVPR16-EnSC} is used to compute the self-expressive coefficients on the entire datasets. Since there is no available GPU acceleration packages for this solver, we run EnSC using an Intel(R) Xeon E5-2630 CPU. The results confirm that our SENet is able to achieve reasonably good performance while using only a small amount of data. This leads to a significantly reduced training time.
For EMNIST, as EnSC needs more than 24 hours, we instead compare SENet to ESC \cite{You:ECCV18} in which 300 exemplars are used. 
Note that SENet achieves comparable performance as ESC within an acceptable time, showing its potential to handle large-scale datasets.

We further compare the performance of SENet to 
other methods in the literature, including $k$-means~\cite{MacQueen-1967}, spectral clustering with normalized cuts (Spectral) \cite{Shi-Malik:PAMI00}, elastic net subspace clustering (EnSC) \cite{You:CVPR16-EnSC}, sparse subspace clustering by orthogonal matching pursuit (SSC-OMP) \cite{You:CVPR16-SSCOMP}, neural collaborative subspace clustering (NCSC) \cite{Zhang:ICML19} and exemplar-based subspace clustering (ESC) \cite{You:ECCV18}. We also compare SENet to several state-of-the-art deep image clustering algorithms, including deep embedded clustering (DEC) \cite{Xie:ICML16}, joint unsupervised learning (JULE) \cite{Yang:CVPR16}, deep adaptive image clustering (DAC) \cite{Chang:ICCV17-DAC}, deep embedded regularized clustering (DEPICT) \cite{Ghasedi:ICCV17-DEPICT}, ClusterGAN \cite{Mukherjee:AAAI19-ClusterGAN}, deep spectral clustering using dual autoencoder network (DSCDAN) \cite{Yang:CVPR19} and deep comprehensive correlation mining (DCCM) \cite{Wu:ICCV19}. 

\vspace{-0pt}
Experimental results are reported in Table \ref{tab:results-on-three datasets compare to SOTA}.
We can see that our SENet is among the best performing methods on the 
four benchmarks.
Specifically, SENet consistently outperforms previous subspace clustering methods on 
CIFAR-10, 
\ie, $+ 15.2\%$ on CIFAR-10 compared to EnSC in terms of accuracy.
Although trained on sampled small datasets, our SENet could still achieve a comparable performance on MNIST-full with significantly reduced training time. Meanwhile, our SENet also achieves comparable performance when compared to state-of-the-art deep image clustering methods. In particular, our SENet outperforms all baseline methods on CIFAR-10 and achieves
second highest accuracy on Fashion-MNIST and EMNIST.

\vspace{-2pt}
\section{Conclusion}
\label{sec:conclusion}
\vspace{-3pt}
We proposed a novel self-expressive network (SENet) for discovering low-dimensional subspace structures in high-dimensional data and presented two stochastic 
gradient descent (SGD) based training algorithms to effectively train SENet.
Different from the conventional self-expressive model, which is defined on the given dataset only and cannot handle out-of-sample data, our proposed SENet is trained on the given dataset and can generalize to unseen new samples.
We conducted extensive experiments on synthetic data and real world data and showed that the self-expressive coefficients learned by SENet are equally good or even better than the self-expressive coefficients learned by a convex self-expressive model.
Moreover, we verified that the out-of-sample ability enables SENet to efficiently handle large-scale dataset.
Beyond the clustering task, self-expressive models also have wide applications in classification \cite{Wright:PAMI09}, exemplar selection \cite{Elhamifar:CVPR12,You:TPAMI20}, outlier/novelty detection \cite{zhang2016sparse,You:CVPR17}, and matrix completion \cite{Elhamifar:NIPS16,Li:TSP16-S3LR,Lane:ICCV2019-missdata} tasks as well, we believe that our SENet may also be extended for many of such tasks, and leave it to future work.

\section*{Acknowledgment}
S. Zhang and C.-G Li are supported by the National Natural Science Foundation of China under Grant 61876022. C.-G. Li is the corresponding author.
R. Vidal was partially supported by the Northrop Grumman Mission Systems Research in Applications for Learning Machines (REALM) initiative, NSF Grants
1704458, 2031985 and 1934979.
C. You acknowledges support from the Tsinghua-Berkeley Shenzhen Institute Research Fund.

{\small
\bibliographystyle{ieee_fullname}
\bibliography{biblio//yc,biblio//cgli,biblio//zhjj_v1,biblio//deeplearning,biblio//sparse,biblio//learning,biblio//vidal,biblio//geometry, biblio//segmentation}
}

\newpage
\appendices


In the appendix, we provide additional experimental results for SENet.
In particular, we provide a set of ablation studies on CIFAR-10 to evaluate the effect of using the learnable soft thresholding activation, varying the number of hidden layers, the number of hidden neurons and the batch size in training.
We provide the learning curves on both synthetic data and real world data.
We also report the running time of SENet on datasets used in our experiments.
Finally, we provide a comparison of Algorithm~\ref{alg:naive} and Algorithm~\ref{alg:two-pass} to demonstrate the scalability of SENet.

\begin{figure*}[!tb]
	\centering
	\subfigure[\small $\L$]{\includegraphics[clip=true,trim=5 8 0 5,width=0.65\columnwidth]{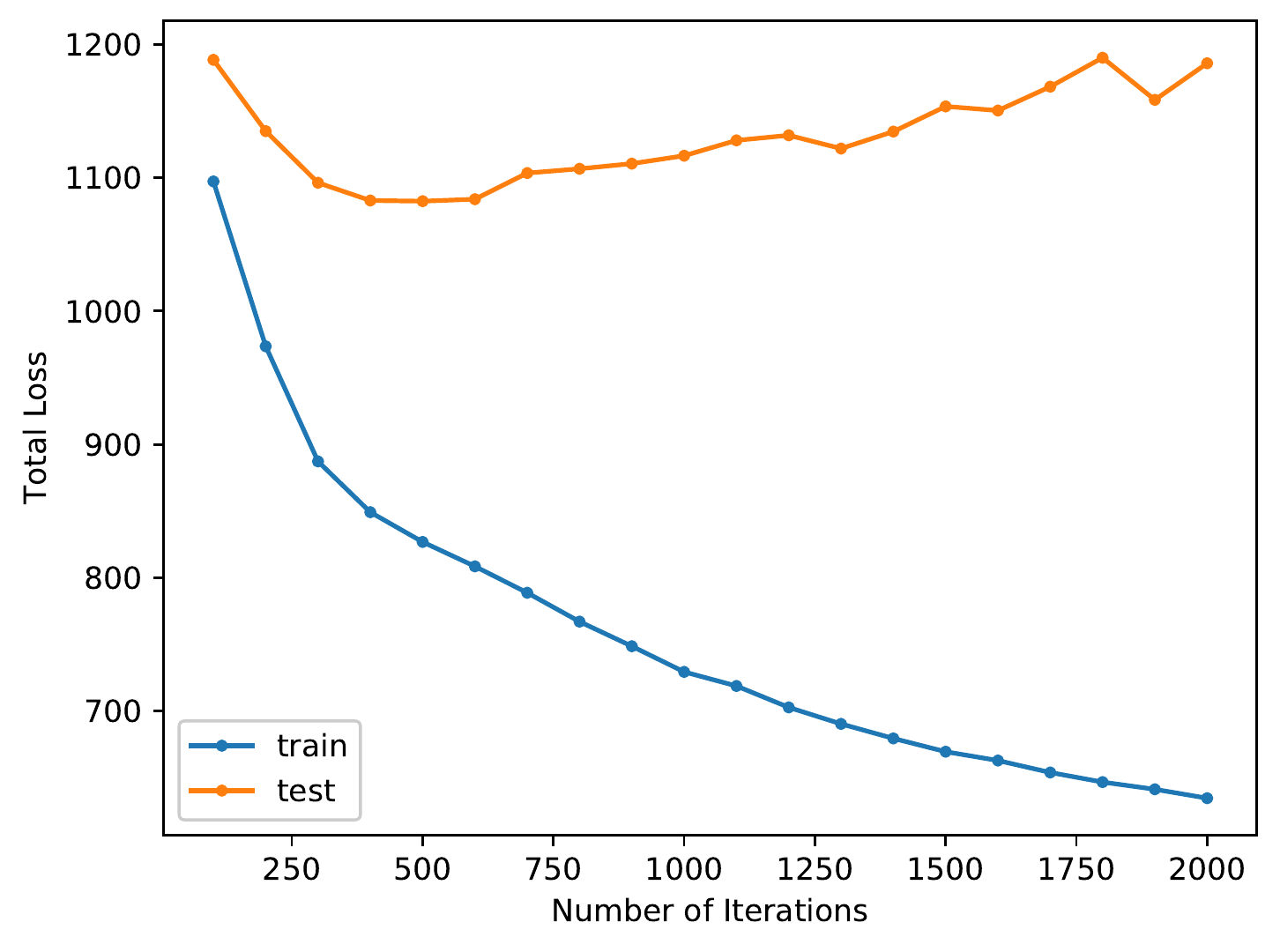}}
	~
	\subfigure[\small $\L_{rec}$]{\includegraphics[clip=true,trim=5 8 0 5, width=0.625\columnwidth]{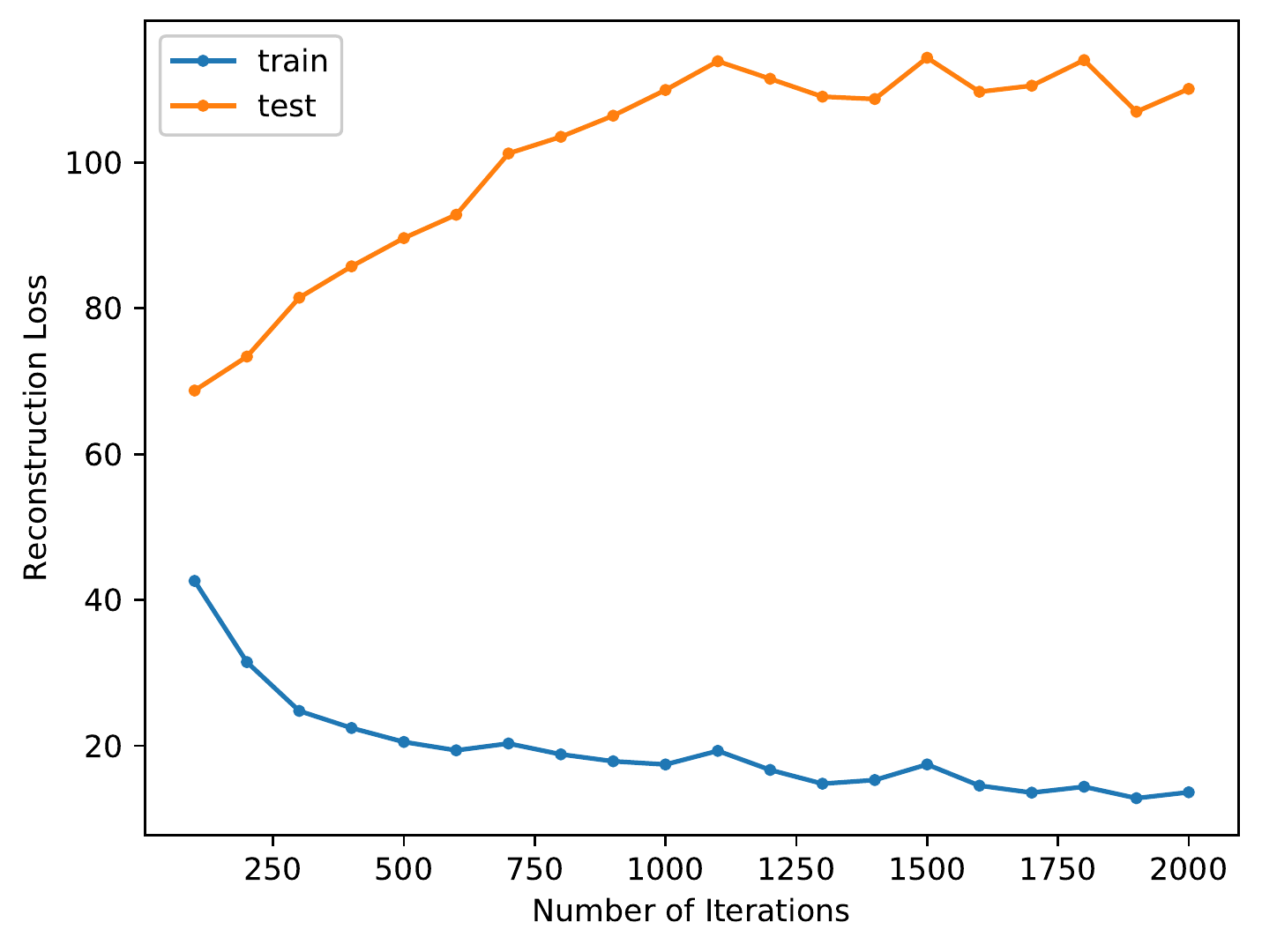}}
	~
	\subfigure[\small $\L_{reg}$]{\includegraphics[clip=true,trim=5 8 0 5, width=0.625\columnwidth]{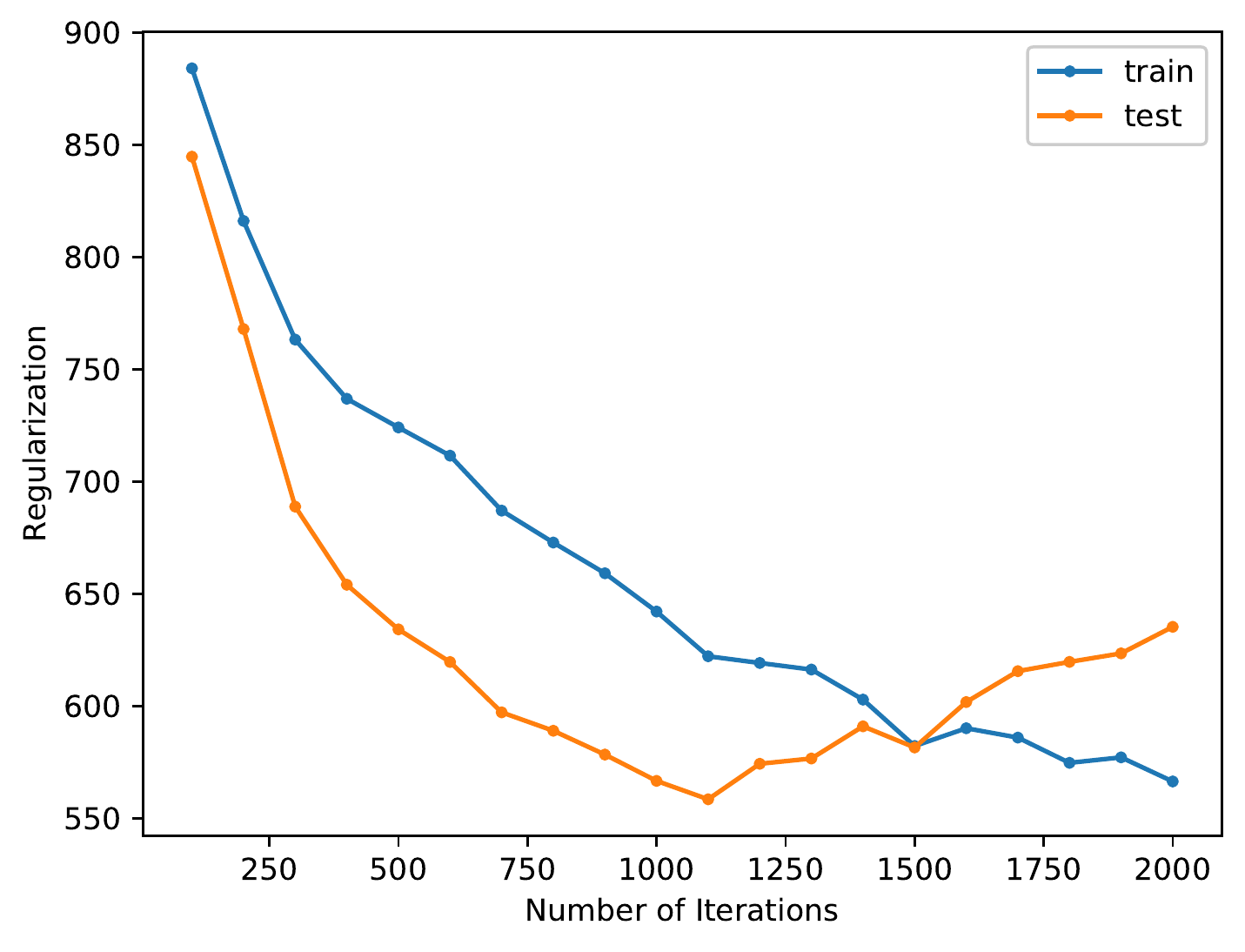}} \\
	\subfigure[\small ACC]{\includegraphics[clip=true,trim=5 8 0 5,width=0.65\columnwidth]{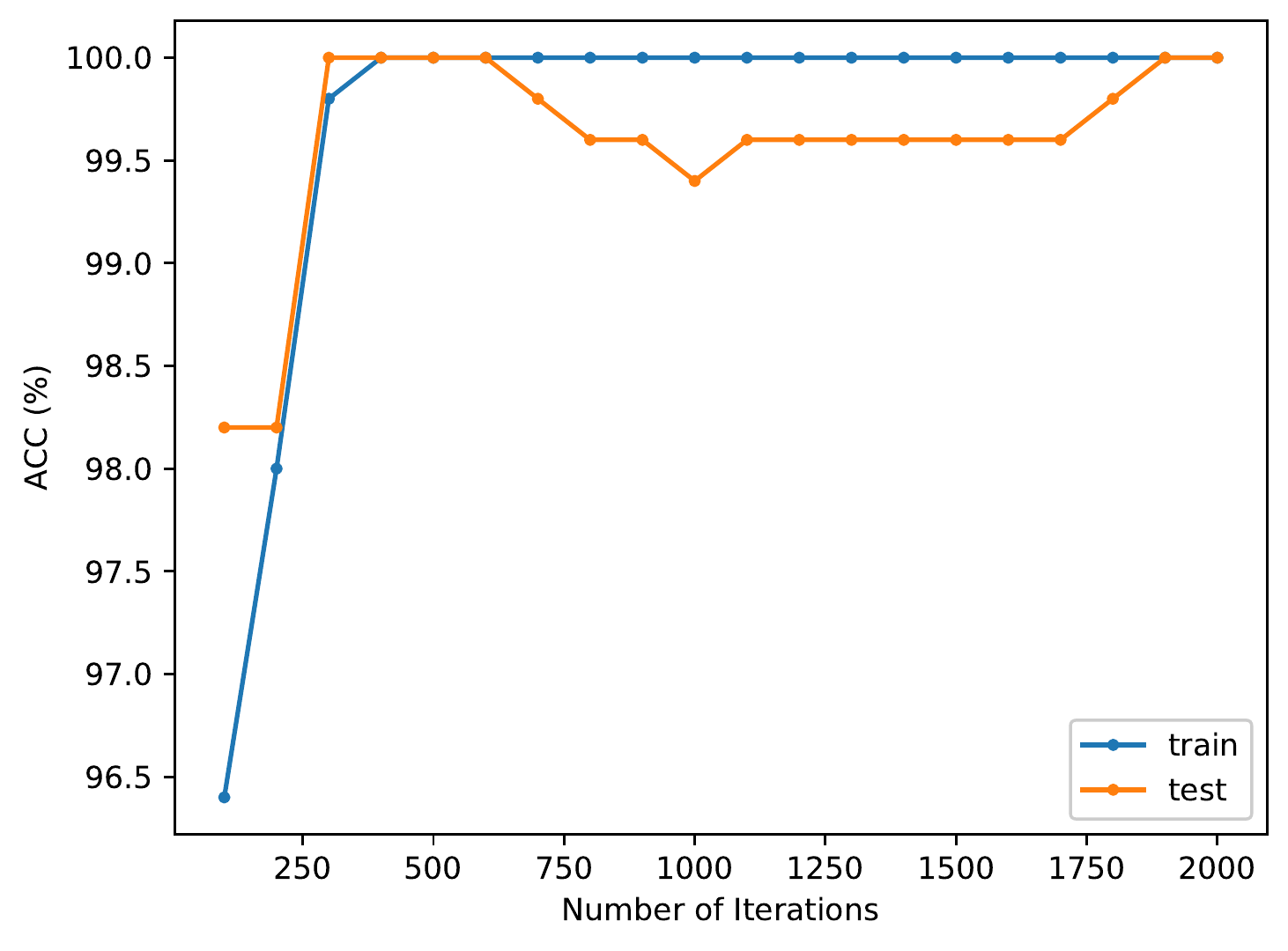}}
	\subfigure[\small SRE]{\includegraphics[clip=true,trim=5 8 0 5, width=0.625\columnwidth]{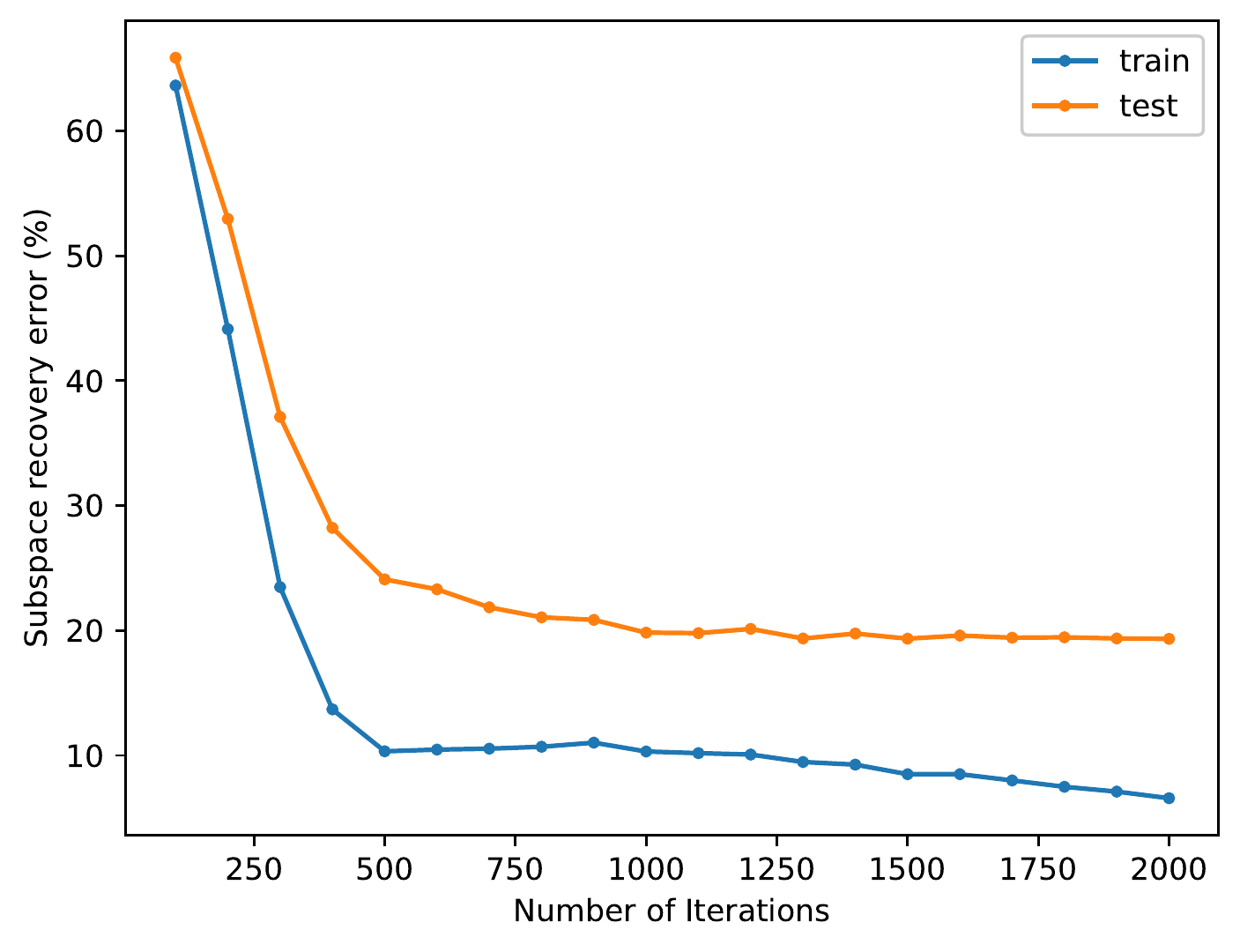}}
	~~~
	\caption{Learning curves for training SENet on the synthetic data (from 100 to 2000 iterations).}
	\label{fig:lr_curves_Synthetic}
\end{figure*}

\section{Additional Results on Synthetic Datasets}
\label{sec:experiments-on-synthetic-data}

\myparagraph{Learning Curves}
We report the losses, ACC and SRE during training, evaluated on both the training and testing data, for the experiments discussed in the paragraph ``Visualization of Self-expressive Coefficients'' of Section \ref{sec:Synthetic Experiments}.
The results are shown in Fig.~\ref{fig:lr_curves_Synthetic}.

We see that on the training data, the loss $\mathcal{L}$ and its two components $\mathcal{L}_\text{rec}$ and $\mathcal{L}_\text{reg}$ all monotonically decrease as training proceeds.
Correspondingly, the ACC monotonically increases and SRE monotonically decreases.
On the other hand, we see that $\mathcal{L}_\text{rec}$ on test data monotonically increases, showing that the SENet does not learn to reconstruct for test data.
Nonetheless, the SRE monotonically decreases on test data, showing that the SENet learns to produce subspace-preserving solutions which leads to an improved ACC. We leave a study of why this behavior occurs to future work.

\section{Additional Results on Real Datasets}
\label{sec:experiments-on-real-data}

\subsection{Effects of Network Design / Training Choices}

We provide additional results for our experiments on clustering 60,000 images in the CIFAR-10 dataset (see Section~\ref{sec:Real Experiments}).
In all experiments below, we train a SENet on $N = $ 2,000 random samples from CIFAR-10 and use it to generate self-expressive coefficients on the entire dataset for clustering.

\myparagraph{SENet with vs. without Soft-Thresholding Layer}
To demonstrate the necessity of the soft-thresholding layer in \eqref{eq:SENet-formulation}, we conduct experiments both with and without the soft-thresholding layer and report results in Table \ref{tab:ablation-soft-thres-CIFAR10}. We can see that the performance of SENet improves significantly with the learnable soft-thresholding layer.

\renewcommand{\arraystretch}{1.1}
\begin{table}[!htbp]
	\centering
	\small
	\begin{tabular}{l|c c c}
		\hline
		{Methods} & ACC & NMI & ARI \\
		\hline
		With soft-thresholding   &0.765	&0.655	&0.573\\
		Without soft-thresholding  &0.610 &0.588 &0.424\\
		\hline
	\end{tabular}
	\\[4pt]
	\caption{Ablation study of soft-thresholding layer on CIFAR-10.}
	\label{tab:ablation-soft-thres-CIFAR10}
\end{table}

\myparagraph{Effect of Network Depth}
We vary the number of layers of the query and key networks in the range of $\{1, 2, 3, 4\}$ while fixing the network width to 1024.
In Table~\ref{tab:ablation-depth-CIFAR10}, we report average and standard deviation for ACC, NMI, ARI and SRE over 5 trials.
We can see that the performance of SENet increases when increasing the number of hidden layers from 1 to 3.
However, further increasing the number of layers beyond three does not help to improve the performance.
Note that even using a single hidden layer, SENet still works well. 

\renewcommand{\arraystretch}{1.1}
\begin{table}[!htbp]
	\centering
	\small
	\scalebox{0.8}{
	\begin{tabular}{c|c c c c}
		\hline
		{Depth} & ACC & NMI & ARI & SRE \\
		\hline
		1	&0.739 $\pm$ 0.002&0.634 $\pm$ 0.001&0.554 $\pm$ 0.002&0.304 $\pm$ 0.001\\
		2	&0.732 $\pm$ 0.022&0.636 $\pm$ 0.012&0.545 $\pm$ 0.016&0.306 $\pm$ 0.001\\
		3	&0.755 $\pm$ 0.016&0.647 $\pm$ 0.009&0.566 $\pm$ 0.016&0.308 $\pm$ 0.001\\
		4	&0.751 $\pm$ 0.013&0.643 $\pm$ 0.008&0.560 $\pm$ 0.011&0.308 $\pm$ 0.001\\
		\hline
	\end{tabular} 	
	}
	\\ [4pt]
	\caption{Effect of network depth on CIFAR-10. We report mean and standard deviation over 5 trials. }
	\label{tab:ablation-depth-CIFAR10}
\end{table}

\begin{figure}[H]
	\centering
	\includegraphics[clip=true,trim=5 8 0 5,width=0.75\columnwidth]{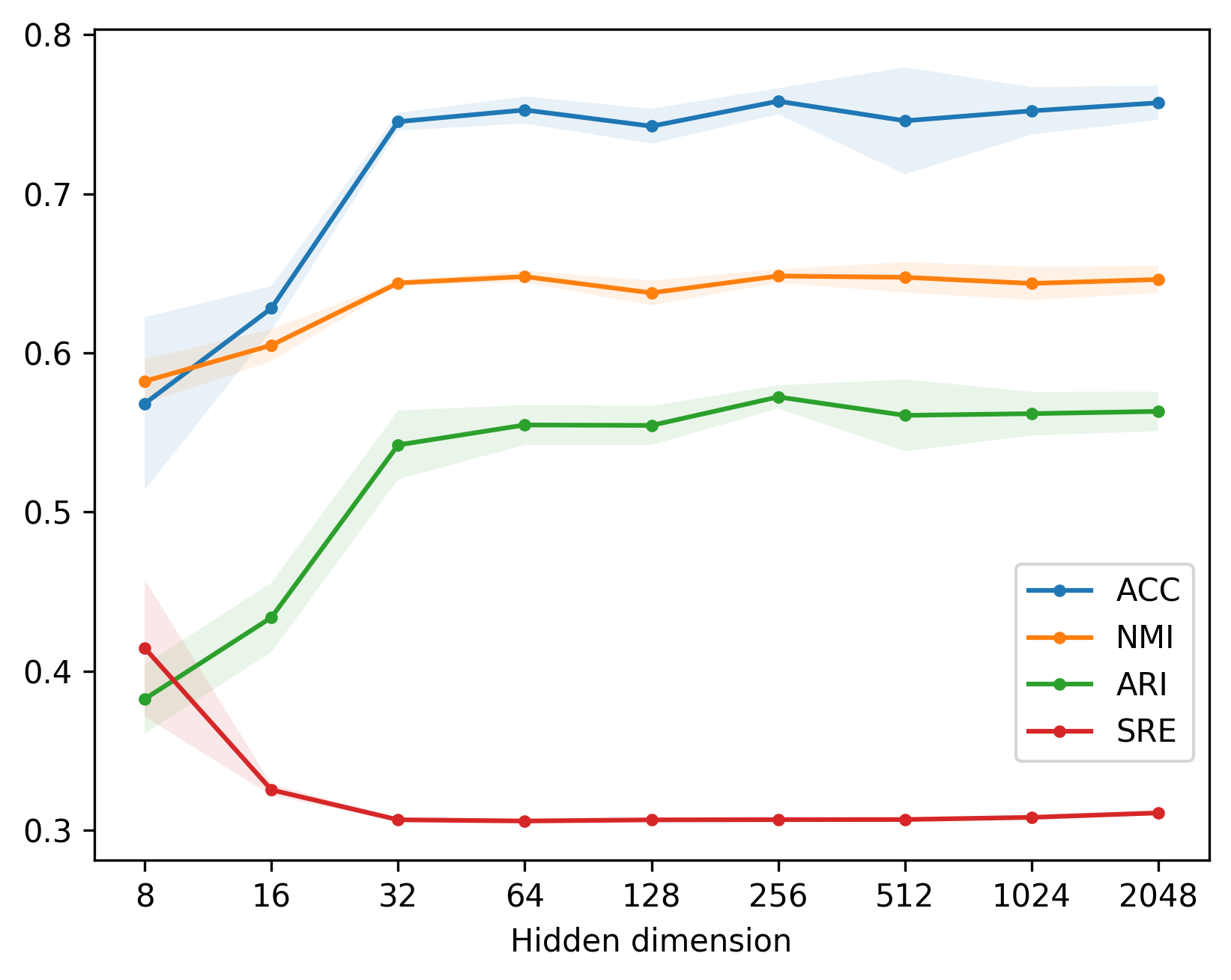}
	\caption{Effect of network width on CIFAR-10. We report mean and standard deviation over 5 trials. }
	\label{fig:ablation-hiddensize-CIFAR10}
\end{figure}

\myparagraph{Effect of Network Width}
We vary the network width (i.e., number of neurons in each hidden as well as output layers of the key and query 
MLPs) in the range of $p \in [8, 2048]$ while fixing the number of layers to be 3.
The parameter $\alpha$ is set to be $1 / p$.
We report the mean (as solid lines) and standard deviation (as shaded areas) for ACC, NMI, ARI and SRE in Fig.~\ref{fig:ablation-hiddensize-CIFAR10}. We can see that the performance continues to improve when the network becomes wider.

\begin{figure}[htbp]
	\centering
	\includegraphics[clip=true,trim=5 8 0 5,width=0.75\columnwidth]{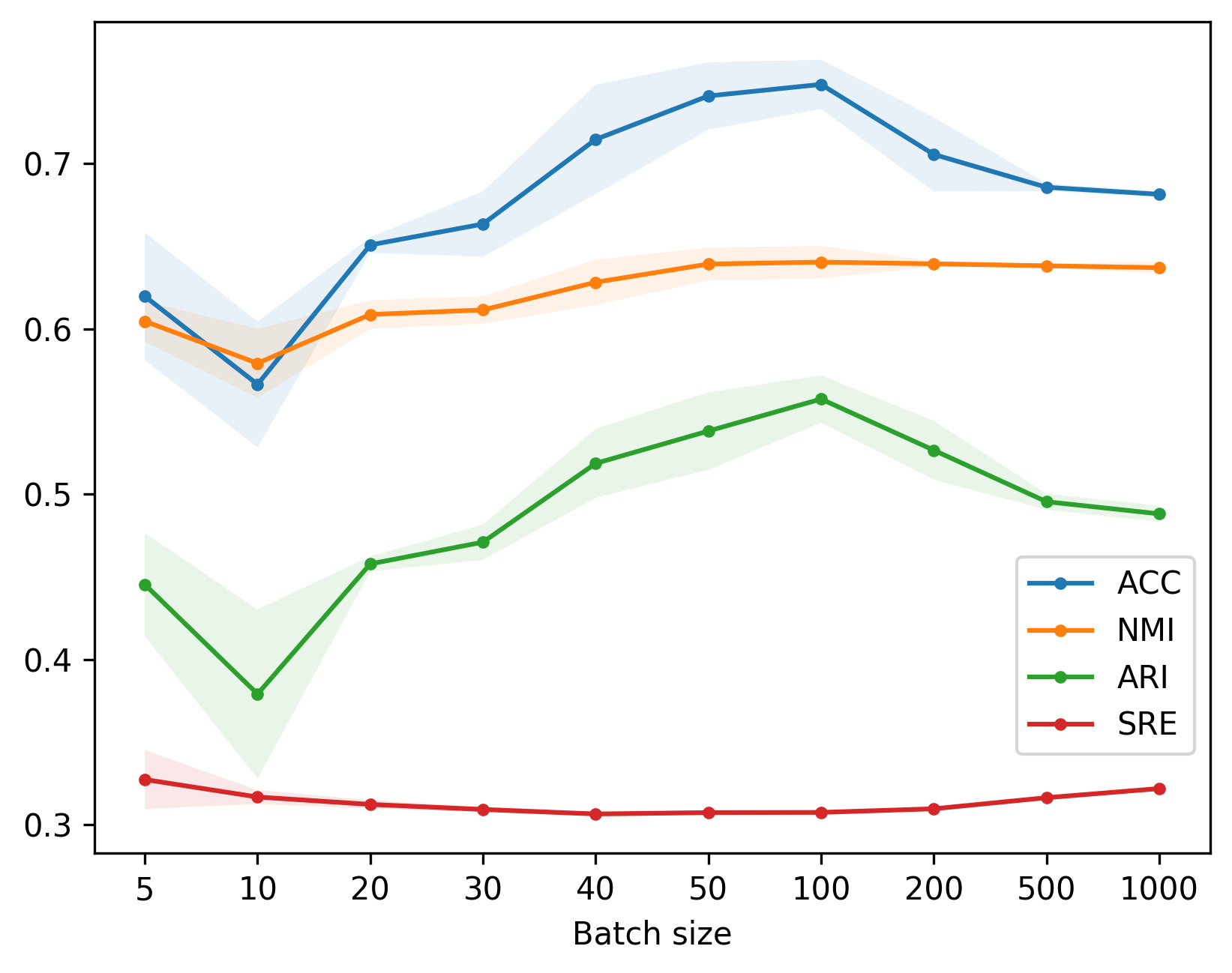}
	\caption{Effect of batch size on CIFAR-10. We report mean and standard deviation over 5 trials. }
	\label{fig:ablation-batchsize-CIFAR10}
\end{figure}

\begin{figure}[htbp]
\centering
\subfigure[Accuracy]{\includegraphics[clip=true,trim=5 8 0 5, width=0.8\columnwidth]{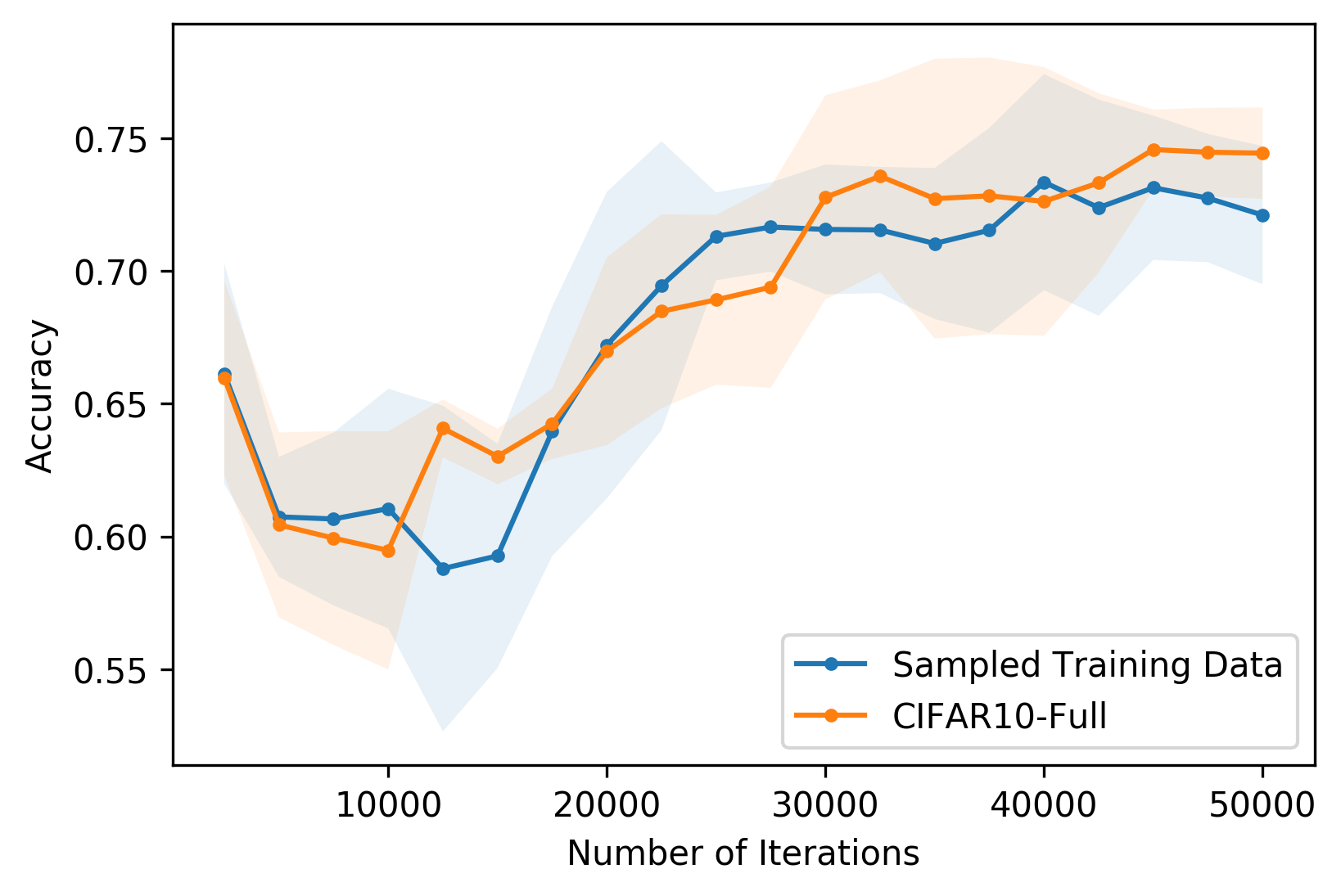}} \\
\subfigure[SRE]{\includegraphics[clip=true,trim=5 8 0 5, width=0.8\columnwidth]{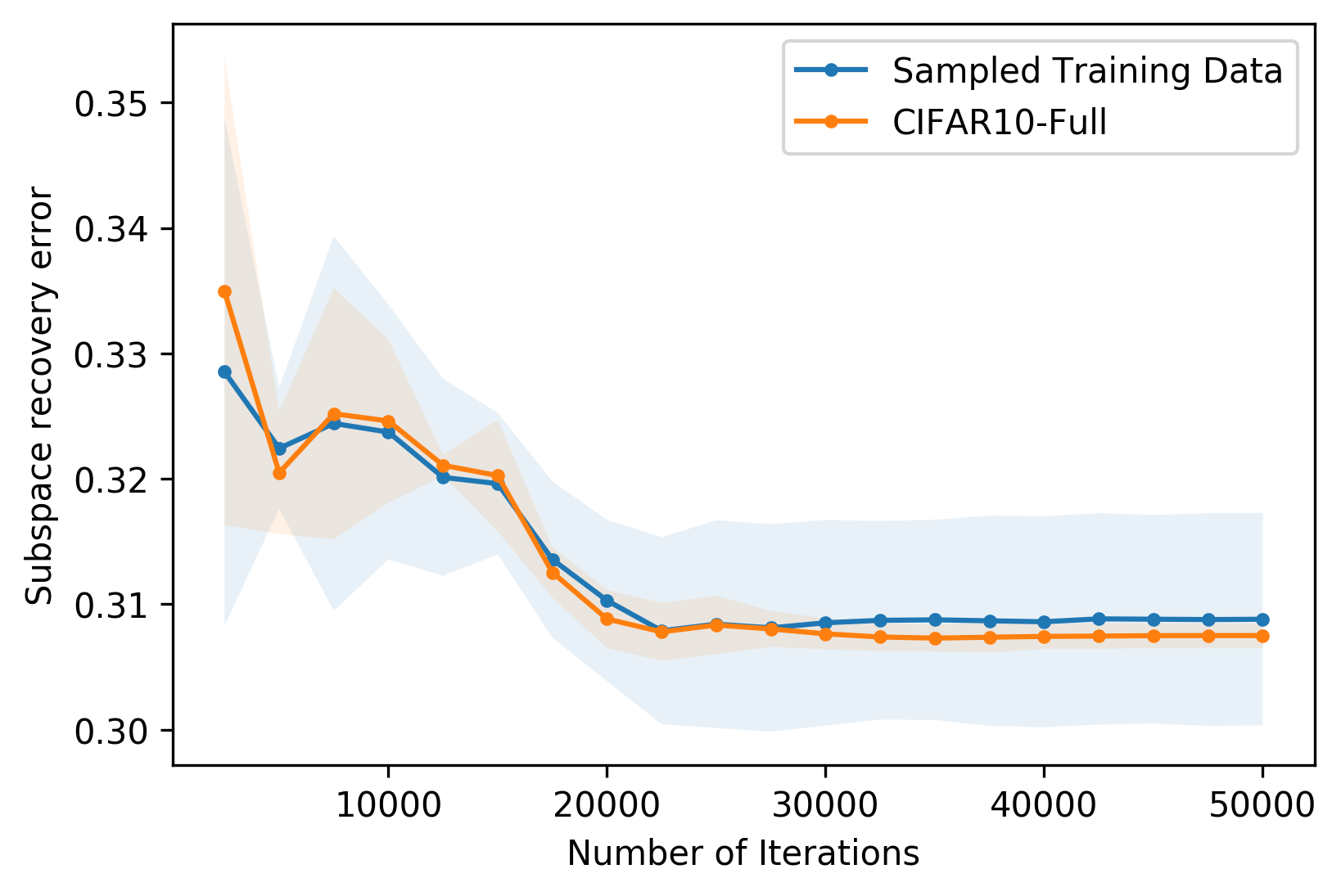}}
\caption{ACC and SRE curves for training SENet on CIFAR-10.}
\label{fig:acc_sre_curves_CIFAR10}
\end{figure}

\myparagraph{Effects of Batch Size}
We evaluate the effect of batch size in the training algorithm and report the mean and standard deviation of ACC, NMI, ARI and SRE in Fig.~\ref{fig:ablation-batchsize-CIFAR10}.
The figure shows that the best batch size should be neither too large nor too small and a batch size of  $\sim$100 produces the best performance in terms of ACC and ARI.

\subsection{Learning Curves}
\begin{figure}[htbp]
	\centering
	\subfigure[\small $\L$]{\includegraphics[clip=true,trim=5 8 0 5,width=0.85\columnwidth]{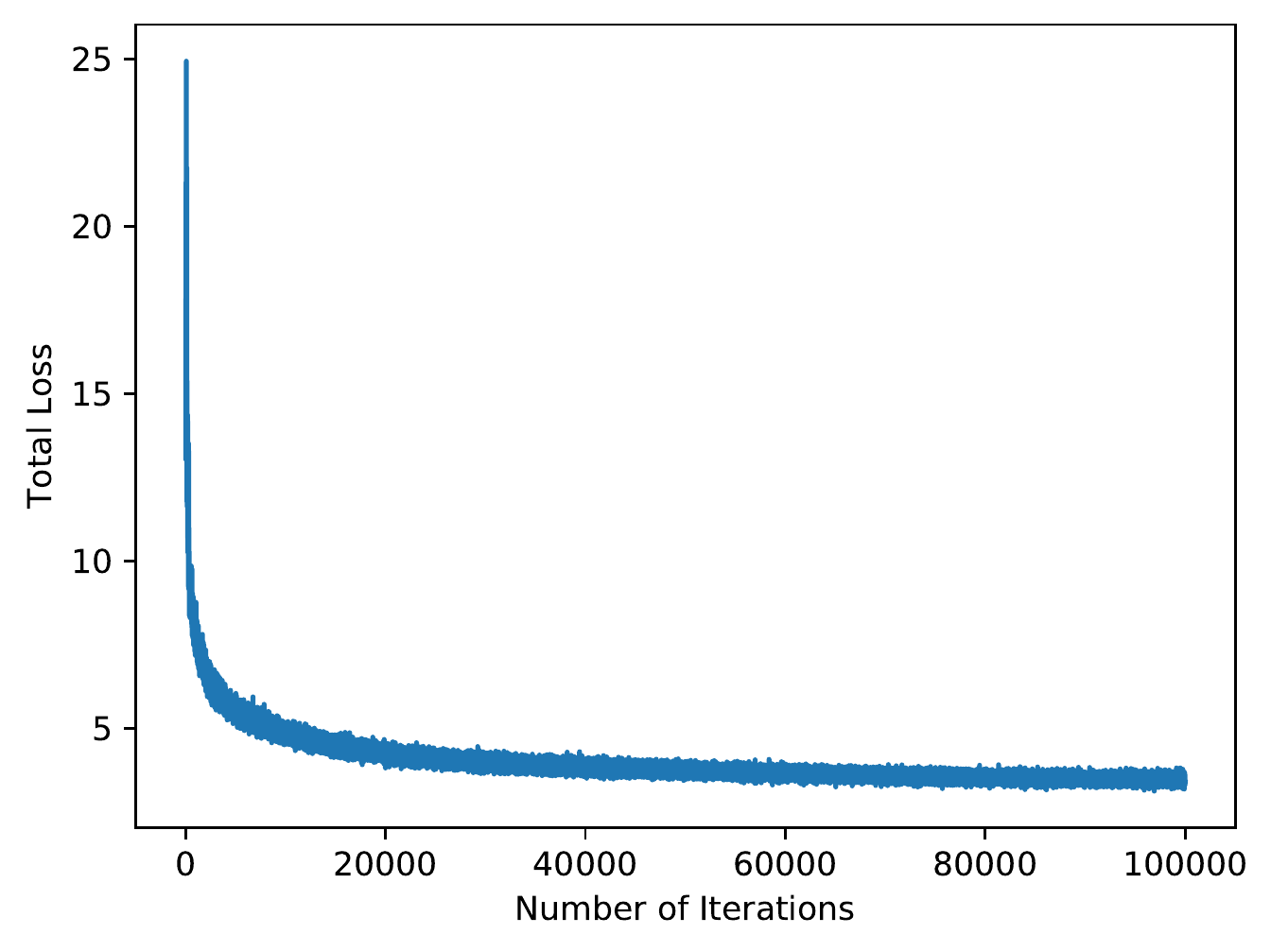}}
	\subfigure[\small $\L_{rec}$]{\includegraphics[clip=true,trim=5 8 0 5, width=0.85\columnwidth]{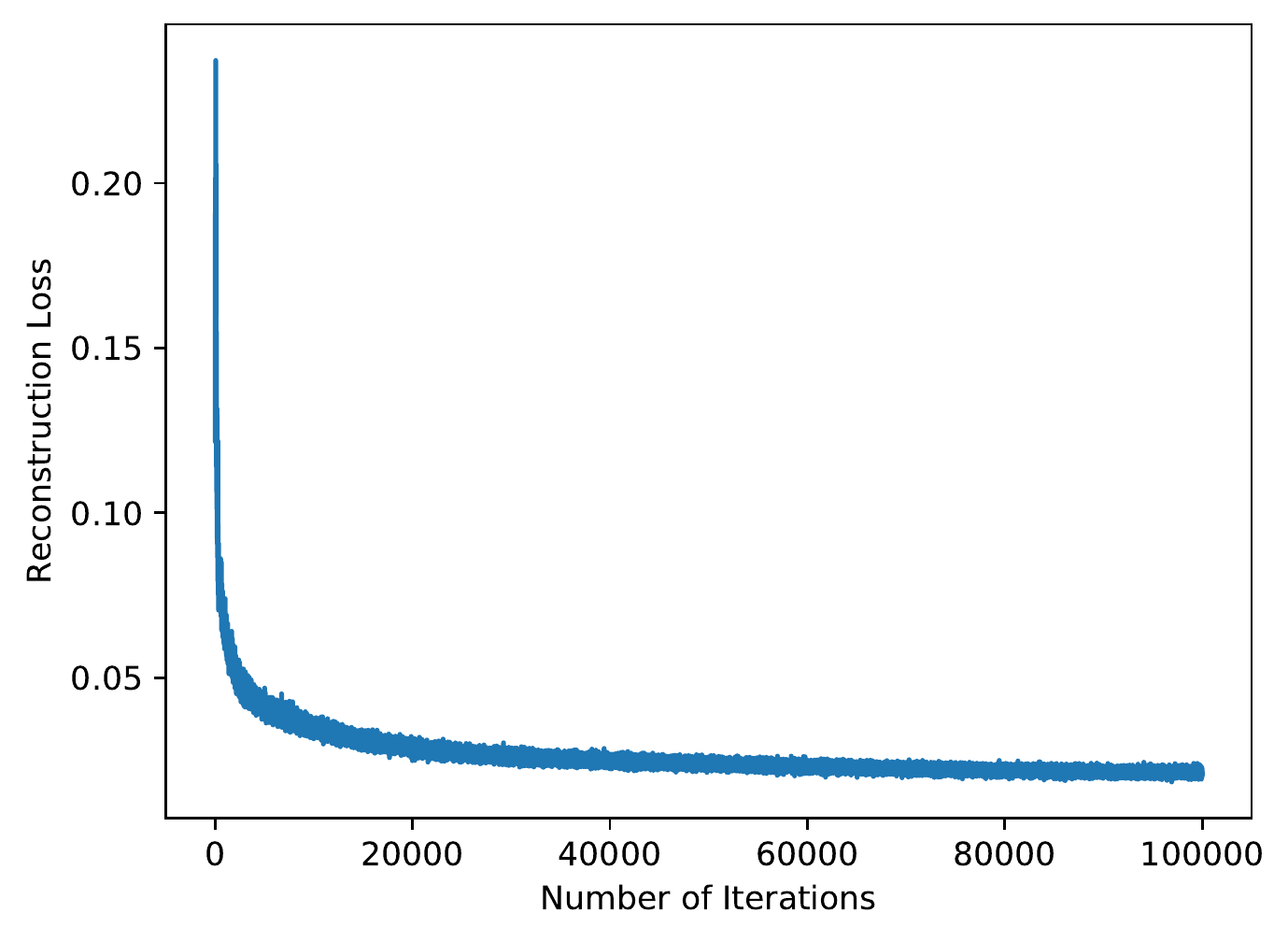}}
	\subfigure[\small $\L_{reg}$]{\includegraphics[clip=true,trim=5 8 0 5, width=0.85\columnwidth]{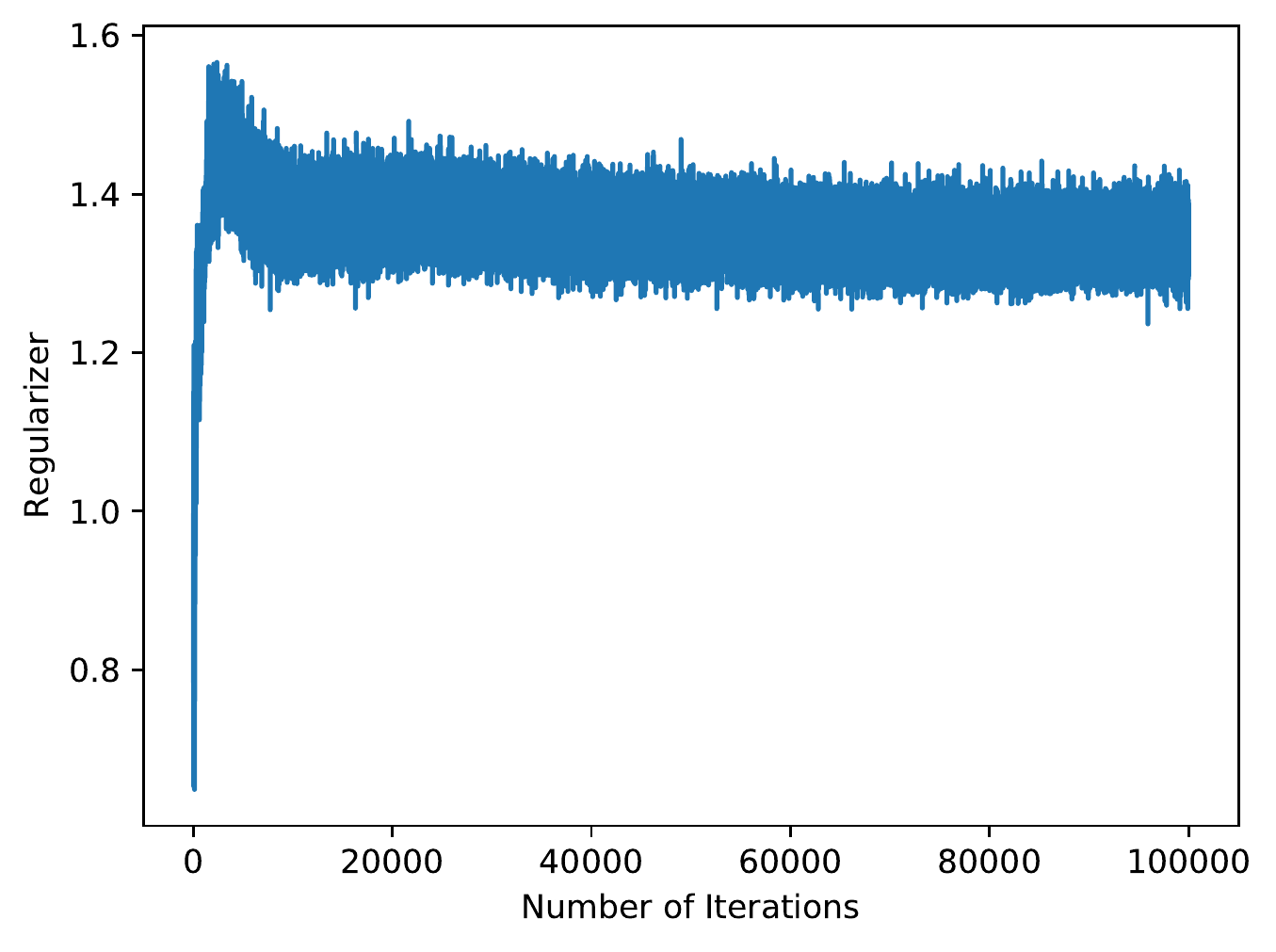}} \\
	\caption{Learning curves for training SENet on MNIST.}
	\label{fig:lr_curves_MNIST}
\end{figure}

\begin{figure*}[htbp]
	\centering
	\subfigure[\small $\L$]{\includegraphics[clip=true,trim=5 8 0 5,width=0.625\columnwidth]{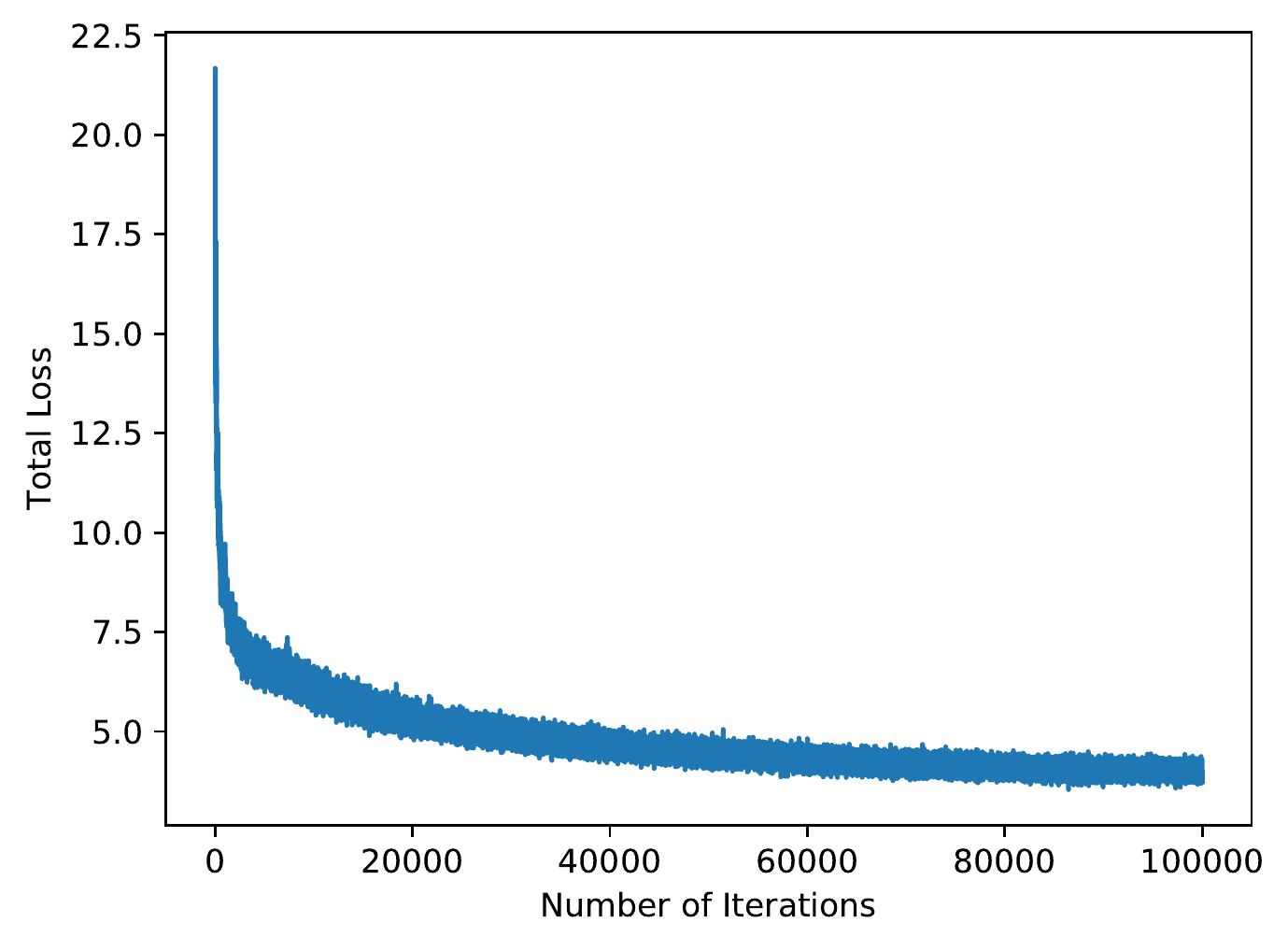}}
	\subfigure[\small $\L_{rec}$]{\includegraphics[clip=true,trim=5 8 0 5, width=0.625\columnwidth]{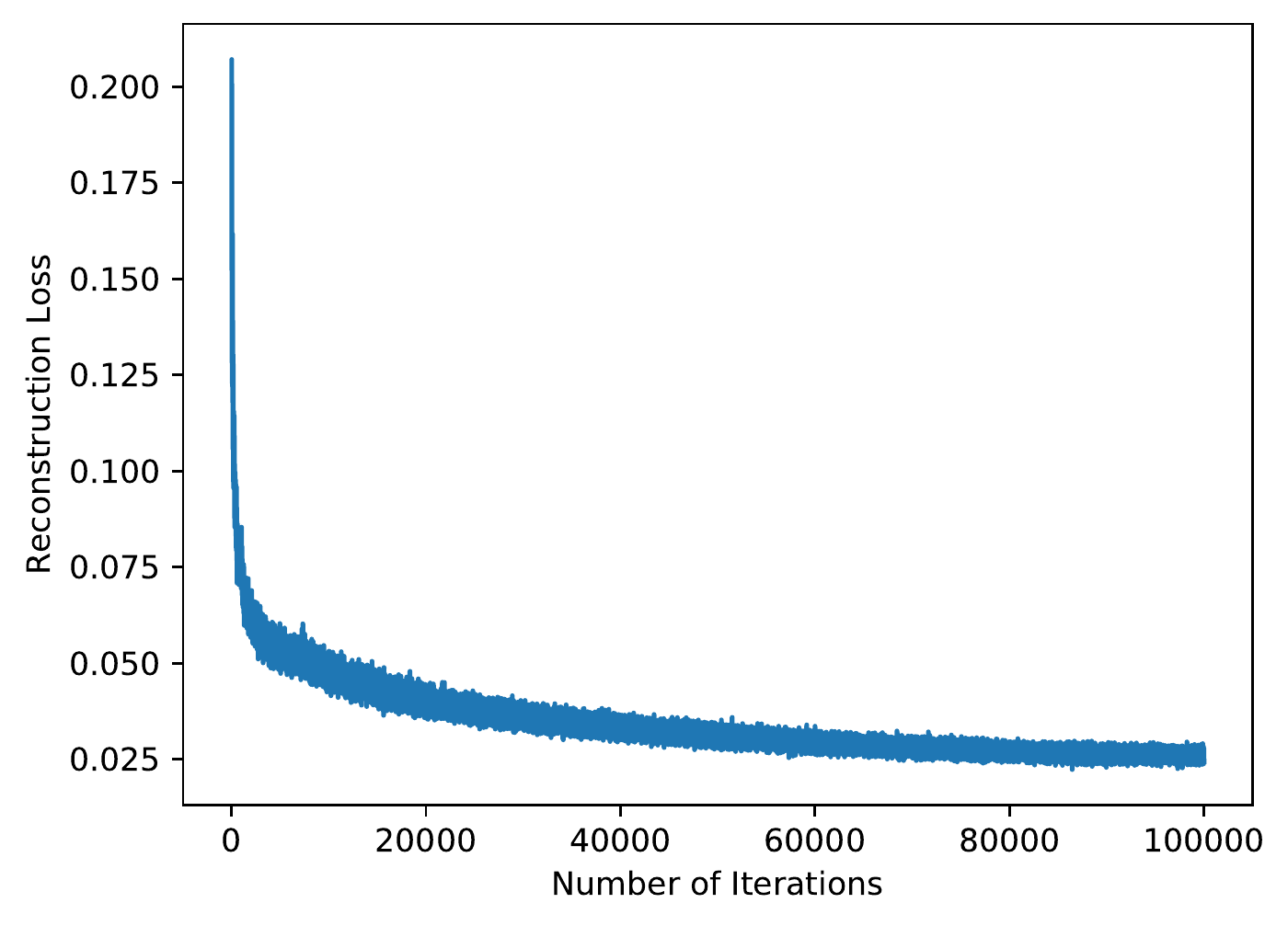}}
	\subfigure[\small $\L_{reg}$]{\includegraphics[clip=true,trim=5 8 0 5, width=0.625\columnwidth]{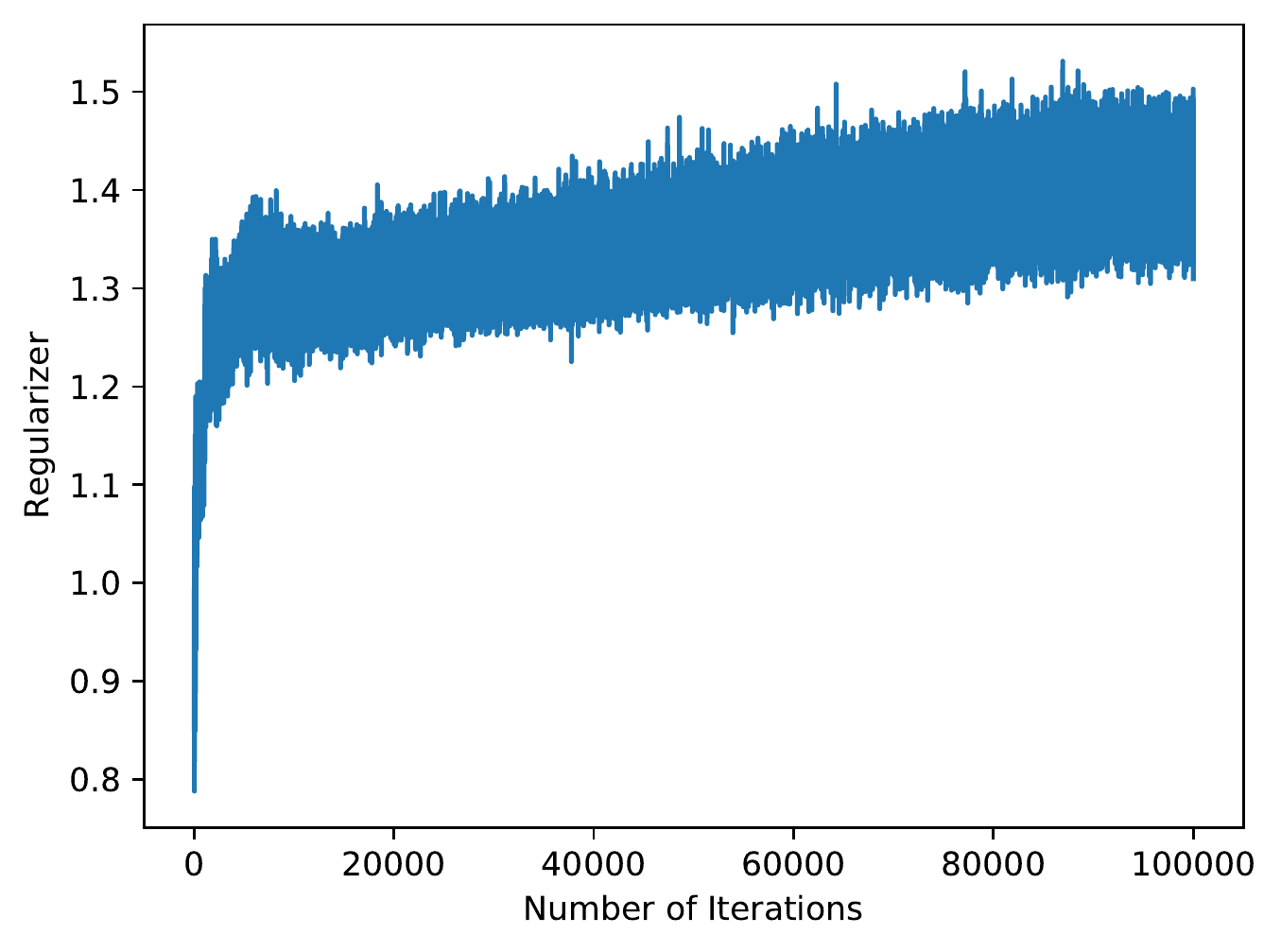}} \\
	\caption{Learning curves for training SENet on FashionMNIST.}
	\label{fig:lr_curves_FashionMNIST}
\end{figure*}

\begin{figure*}[htbp]
	\centering
	\subfigure[\small $\L$]{\includegraphics[clip=true,trim=5 8 0 5,width=0.625\columnwidth]{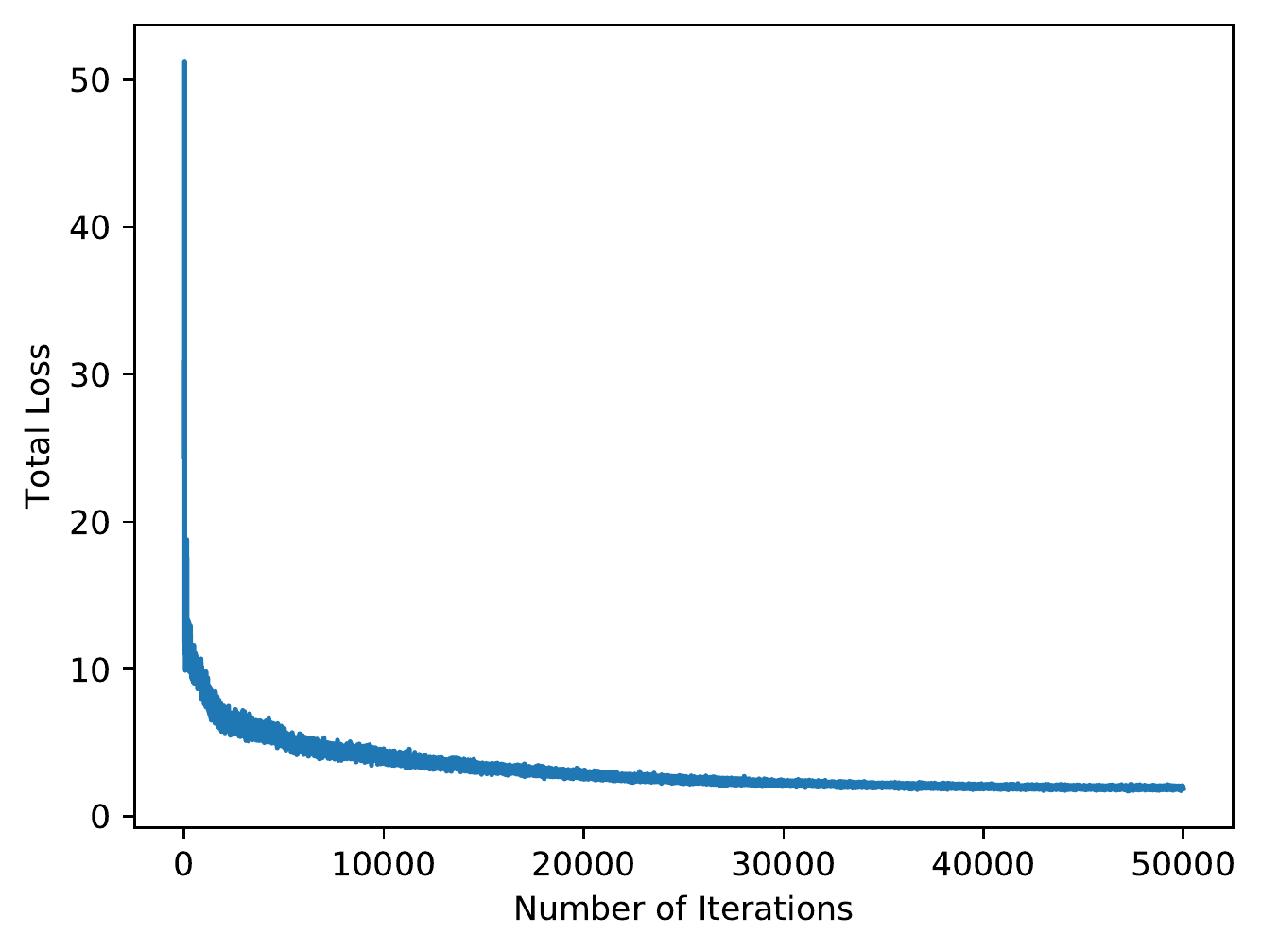}}
	\subfigure[\small $\L_{rec}$]{\includegraphics[clip=true,trim=5 8 0 5, width=0.625\columnwidth]{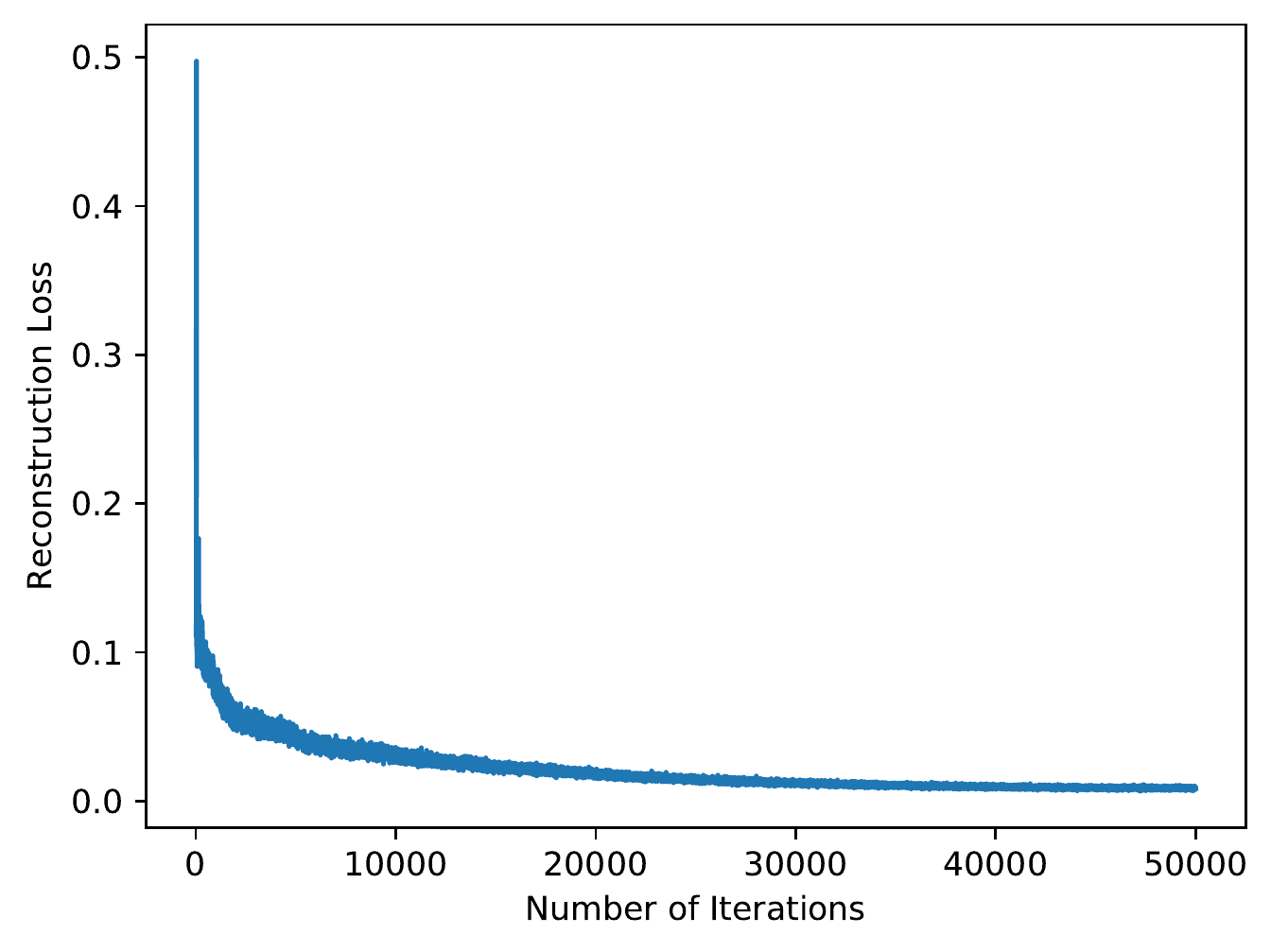}}
	\subfigure[\small $\L_{reg}$]{\includegraphics[clip=true,trim=5 8 0 5, width=0.625\columnwidth]{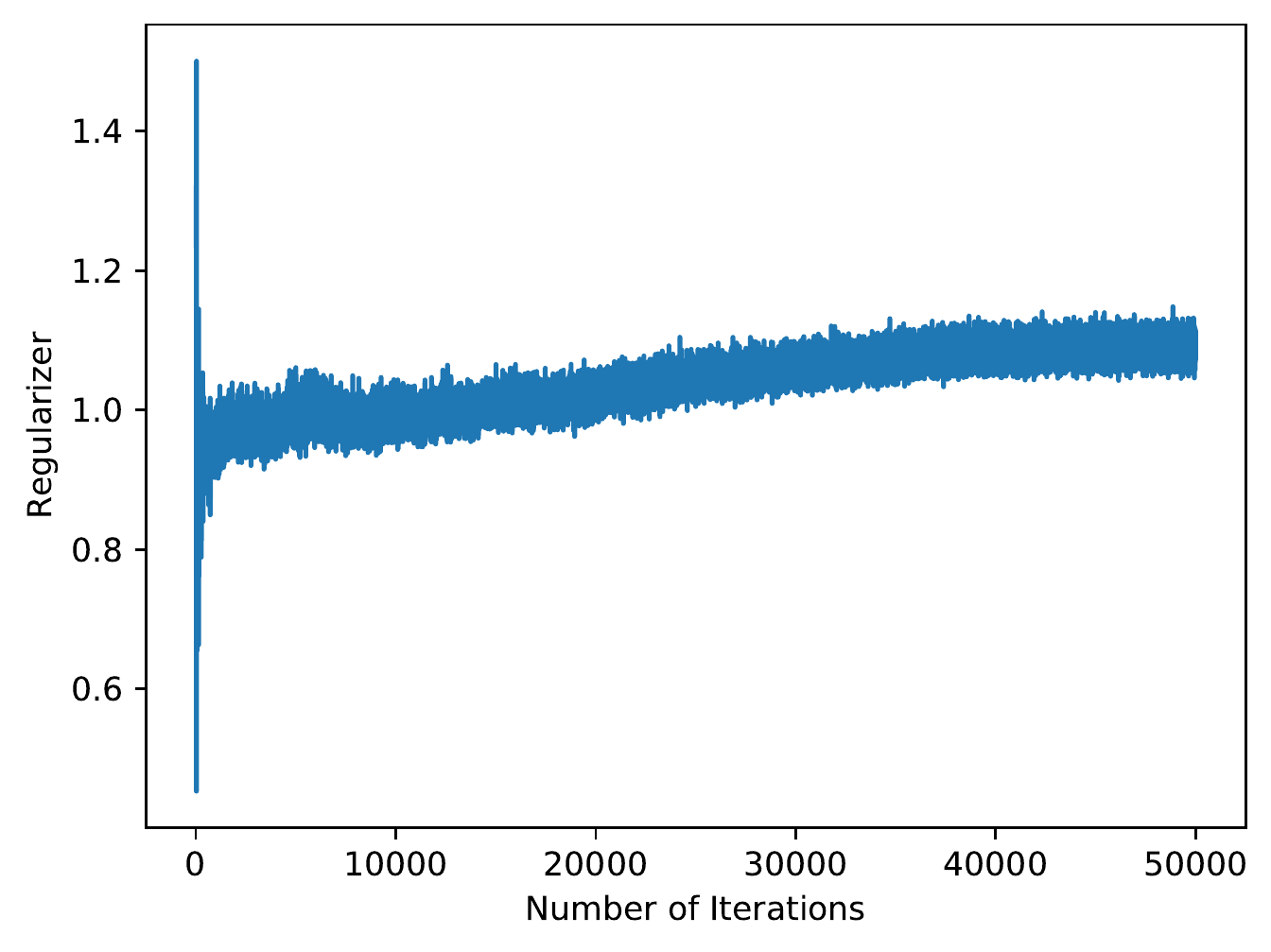}} \\
	\caption{Learning curves for training SENet on CIFAR-10.}
	\label{fig:lr_curves_CIFAR10}
\end{figure*}

We plot the curves of the total loss $\L$, the reconstruction loss $\L_{rec}$ and the regularization term $\L_{reg}$ during training for experiments on the MNIST, Fashion-MNIST and CIFAR-10 datasets as discussed in Section~\ref{sec:Real Experiments}.
The number of training data is fixed to 2,000.
%
The results are shown in Fig.~\ref{fig:lr_curves_MNIST},  Fig.~\ref{fig:lr_curves_FashionMNIST} and  Fig.~\ref{fig:lr_curves_CIFAR10}.
We can observe that 
the total loss $\L$ and the reconstruction loss $\L_{rec}$ are steadily decreasing during training on all the three datasets. Nonetheless, the trends of the regularizer term $\L_{reg}$ are slightly different in these three datasets.  The $\L_{reg}$ term shows a rapid growth at the early stage of the training in all three datasets but it tends to slightly increase on FashionMNIST and CIFAR-10 and slightly decrease on MNIST.


For CIFAR-10, we further plot ACC and SRE on the sampled training data (containing 2,000 data points), as well as on the entire CIFAR-10 dataset (containing 60,000 data points).
The mean and standard deviation of the results computed over 5 trials are reported in Fig.~\ref{fig:acc_sre_curves_CIFAR10}.
We can see that the ACC increases with the number of iterations while the SRE decreases first and then stabilizes.

\subsection{Time Evaluation in Detail}
\label{sec:time-evaluation}


Recall that Fig.~\ref{fig:Real-Dataset-Diff-DictSize} reports the running time of SENet for clustering MNIST, Fashion-MNIST and CIFAR-10 datasests.
The running time can be decomposed into two parts: the first part is the training time for SENet on the subsampled training set, and the second part is the inference time for computing the self-expressive coefficients on the entire dataset with a trained SENet.

We report the training and inference time with varying sizes for the training set in Table \ref{tab:time-on-three datasets}.
The result demonstrates that the inference speed of SENet is very fast and it also has a reasonably small training time. Together, 
this enables SENet to efficiently handle large-scale clustering problems.

\renewcommand{\arraystretch}{1.1}
\begin{table*}[!b]
	\centering
	\small
	\begin{tabular}{l|l|c c c|c c c|c c c}
		\hline
		\multirow{2}{*}{Methods} & \multirow{2}{*}{ Train Data } &\multicolumn{3}{|c|}{MNIST}          &\multicolumn{3}{|c}{Fashion-MNIST}
		&\multicolumn{3}{|c}{CIFAR-10} \\
		\multirow{2}{*}{} & & Training & Inference & Total
		& Training & Inference & Total
		& Training & Inference & Total \\
		\hline
		EnSC    & -    &-    &-   &10252.71
							&-    &-   &10583.55
							&-    &-   &2237.14\\

		\hline
		\multirow{7}{*}{SENet}
		& 200       &1164.63	&300.37	&1465.00
					&1081.53	&299.83	&1381.36
					&499.31	    &13.53	&512.85\\
		& 500       &1070.80    &276.20	&1347.00
					&1180.33    &304.13	&1484.45
					&497.24	    &12.57	&509.81\\
		& 1000      &1213.36	&271.89	&1485.25
				    &1102.98	&279.98	&1382.95
					&510.09	    &19.59	&529.67\\
		& 2000      &1423.43	&254.59	&1678.01
					&1384.23	&310.21	&1694.44
					&617.43	    &19.32	&636.75\\
		& 5000      &2331.42	&267.80	&2599.21
					&2277.91	&330.74	&2608.64
					&892.46	    &21.79	&914.24\\
		& 10000     &3921.22    &282.14	&4203.35
					&3865.35	&329.59	&4194.95
					&1509.32	&19.27	&1528.59\\
		& 20000     &7316.09	&285.82	&7601.91
					&7161.92	&343.26	&7505.19
				    &2982.29	&20.38	&3002.66\\
		\hline

	\end{tabular}
	\\[4pt] 	
	\caption{Training time (sec.) on MNIST-full, Fashion-MNIST-full and CIFAR-10.}
	\label{tab:time-on-three datasets}
\end{table*}

\subsection{Comparison of Algorithm~\ref{alg:naive} and Algorithm~\ref{alg:two-pass}}
\label{sec:time-evaluation}

In all aforementioned experiments, we train SENet with Algorithm~\ref{alg:naive}.
As discussed in Section~\ref{subsec:training}, the memory consumption of Algorithm~\ref{alg:naive} grows with the training data size, preventing its application to large-scale datasets.
In contrast, Algorithm~\ref{alg:two-pass}, while being equivalent to  Algorithm~\ref{alg:naive}, has a fixed memory consumption and in principle can handle arbitrarily large datasets.
While memory requirement does not become an issue for experiments in this paper, this section provides additional experiments to validate the advantage of Algorithm~\ref{alg:two-pass} over Algorithm~\ref{alg:naive}, hence demonstrates the promise of SENet in handling data with larger scale than those used in this paper.

To illustrate the effect of training data size, we conduct experiments with $N \in \{1000, 2000, 5000, 10000, 20000\}$ data points sampled from MNIST-train. We set batch size (i.e., the number of samples in step~\ref{step:batch-sample} of Algorithm~\ref{alg:naive} and step~\ref{step:two-pass-batch-sample} of Algorithm~\ref{alg:two-pass}) to 100 in both algorithms.
While the for-loops at both step~\ref{step:two-pass-first-iterate} and step~\ref{step:two-pass-second-iterate} of Algorithm~\ref{alg:two-pass} process one data point at each iteration, they can both be parallelized into processing multiple data points (referred to as a block) to reduce computational overheads hence accelerate the algorithm.
For simplicity of presentation we fix the block size to be 1000 in our experiments, but point out that block size controls a trade-off between computation efficiency and memory requirement hence needs to be chosen according to the availability of computational resources in practice.

\begin{figure*}[htbp]
	\centering
	\subfigure[Time per iteration]{\includegraphics[clip=true,trim=5 8 0 5, width=0.65\columnwidth]{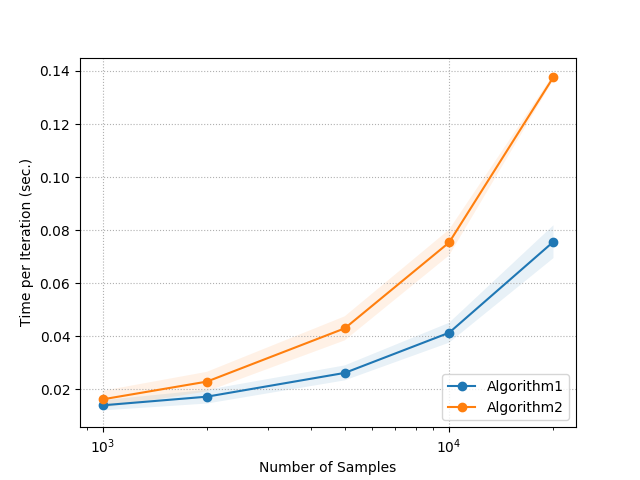}}
	\subfigure[GPU Memory Usage]{\includegraphics[clip=true,trim=5 8 0 5, width=0.65\columnwidth]{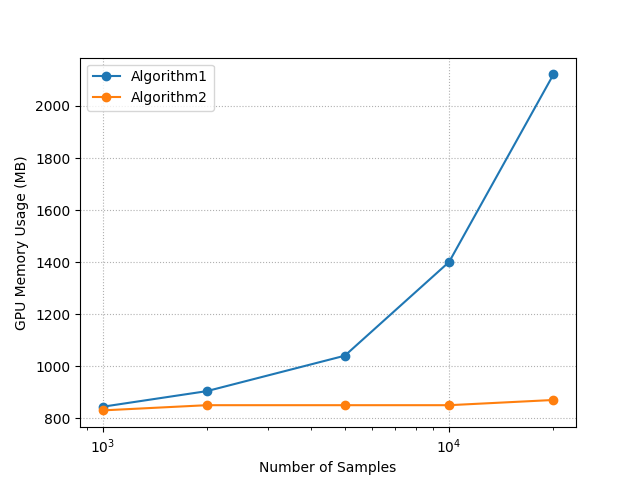}}
	\subfigure[Clustering Accuracy]{\includegraphics[clip=true,trim=5 8 0 5, width=0.65\columnwidth]{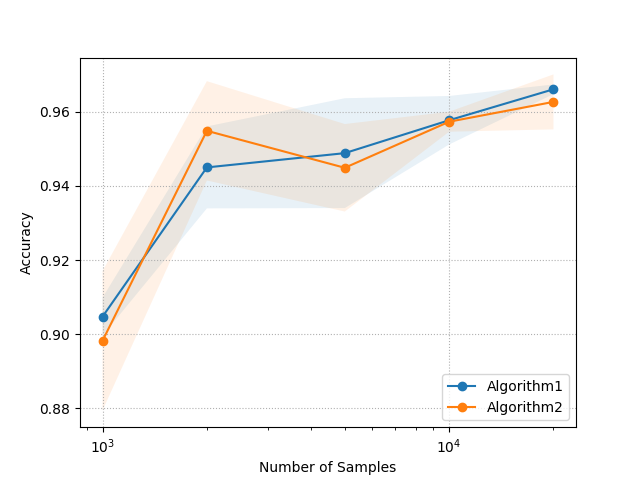}}
	\caption{Comparison of Algorithm~\ref{alg:naive} and Algorithm~\ref{alg:two-pass} with varying dataset size on MNIST.}
	\label{fig:cmp_alg1_alg2}
\end{figure*}

We conduct 4 independent trials with Algorithm~\ref{alg:naive} and Algorithm~\ref{alg:two-pass} and report running time per iteration, the GPU memory usage and clustering accuracy in Fig.~\ref{fig:cmp_alg1_alg2}.
We can observe that the GPU memory consumption of Algorithm~\ref{alg:two-pass} does not change with the size of training data.
Meanwhile, Algorithm~\ref{alg:two-pass} is slightly slower than Algorithm~\ref{alg:naive} since 1) it has the overhead of processing blocks at step~\ref{step:two-pass-first-iterate} and step~\ref{step:two-pass-second-iterate} rather than processing all data at once (as in Algorithm~\ref{alg:naive}), and 2) it takes an additional forward pass.
Finally, both algorithms yield almost the same clustering accuracy, hence verifying their equivalence.

\end{document}